\documentclass[journal]{IEEEtran}

\usepackage{lscape}

\usepackage{longtable}

\usepackage{url}
\usepackage[font=footnotesize]{caption}
\usepackage{subcaption}
\usepackage{comment}
\usepackage{enumitem}

\usepackage[normalem]{ulem}
\renewcommand\sout[1]{}%

\usepackage[tableposition=top]{caption}

\usepackage{amsmath,amssymb,amsfonts}
\usepackage{bm}
\usepackage{algorithm}
\usepackage[noend]{algpseudocode}
\usepackage{graphicx}
\usepackage[table, svgnames]{xcolor}
\usepackage[export]{adjustbox}

\usepackage[usenames,dvipsnames]{pstricks}
\usepackage[dvipsnames]{xcolor}
\usepackage{epsfig}
\usepackage{amssymb}
\usepackage{amsmath}
\usepackage{lscape}
\usepackage{makecell}
\usepackage{longtable}
\usepackage{soul}
\usepackage[colorlinks = true,
  linkcolor  = mylinkcolor,
  citecolor  = mycitecolor,
  urlcolor   = Black ]{hyperref}
\usepackage[numbers,sort&compress]{natbib}
\usepackage{balance}

\usepackage{xcolor,pifont}
\newcommand*\colourcheck[1]{%
  \expandafter\newcommand\csname #1check\endcsname{\textcolor{#1}{\ding{52}}}%
}
\colourcheck{blue}
\colourcheck{green}
\colourcheck{ForestGreen}
\usepackage{duckuments} 

\usepackage{multirow}
\usepackage{ragged2e}
\usepackage{tabularx}
\usepackage[capitalise]{cleveref}
\usepackage{ulem}
\usepackage{amsthm}

\usepackage{titlesec}
\titlespacing\subsection{0pt}{4pt plus 2pt minus 2pt}{1pt plus 2pt minus 2pt}
\titlespacing\subsubsection{0pt}{4pt plus 2pt minus 2pt}{1pt plus 2pt minus 2pt}

\usepackage[subtle]{savetrees}

\colorlet{mylinkcolor}{Black}
\colorlet{mycitecolor}{Red}
\colorlet{myurlcolor}{Magenta}
\newcolumntype{?}{!{\vrule width 1.5pt}}
\hypersetup{
  colorlinks = true,
  linkcolor  = mylinkcolor,
  citecolor  = mycitecolor,
  urlcolor   = cyan
}
\usepackage{colortbl}

\definecolor{gray_table}{HTML}{f0f0f5}

\usepackage{tikz}
\usetikzlibrary{mindmap}
\usetikzlibrary{external}
\usepackage{pgfplots}
\pgfplotsset{compat=1.9}
\usepgfplotslibrary{statistics}
\usepgfplotslibrary{fillbetween}

\title{\LARGE \bf Deep Learning Approaches to Grasp Synthesis: A Review}

\begin{document}

\rowcolors{1}{White}{gray_table}

\author{Rhys Newbury$^{1,2}$,
        Morris Gu$^{1}$,
        Lachlan Chumbley$^{1}$,
        Arsalan Mousavian$^{3}$, Clemens Eppner$^{3}$, 
        J\"{u}rgen Leitner$^{1,4}$,
        Jeannette Bohg$^{5}$, Antonio Morales$^{6}$, Tamim Asfour$^{7}$, Danica Kragic$^{8}$,
        Dieter Fox$^{3, 9}$, Akansel Cosgun$^{10}$
\thanks{\hspace*{-1em}$^{1}$  Monash University, Australia\\$^{2}$  The Australian National University, Australia\\$^{3}$  NVIDIA Corporation, USA\\$^{4}$  LYRO Robotics Pty Ltd, Australia\\$^{5}$  Stanford University, USA\\$^{6}$  Jaume I University, Spain\\$^{7}$  Karlsruhe Institute of Technology, Germany\\$^{8}$  KTH Royal Institute of Technology, Sweden\\$^{9}$  University of Washington, USA\\$^{10}$ Deakin University, Australia}
}
\maketitle

\begin{abstract}


Grasping is the process of picking up an object by applying forces and torques at a set of contacts. Recent advances in deep-learning methods have allowed rapid progress in robotic object grasping. In this systematic review, we surveyed the publications over the last decade, with a particular interest in grasping an object using all 6 degrees of freedom of the end-effector pose. Our review found four common methodologies for robotic grasping: sampling-based approaches, direct regression, reinforcement learning, and exemplar approaches. Additionally, we found two `supporting methods` around grasping that use deep-learning to support the grasping process, shape approximation, and affordances. We have distilled the publications found in this systematic review (85 papers) into ten key takeaways we consider crucial for future robotic grasping and manipulation research. An online version of the survey is available at \url{https://rhys-newbury.github.io/projects/6dof/}


\end{abstract}

\section{Introduction}\label{sec:intro}

The manipulation of objects is an essential skill for robot in the real-world, with grasping being an integral part of such tasks. Grasping is the process of controllings an object's motion in a desired way by applying forces and torques at a set of contacts. It is an essential ability required for the majority of object manipulation tasks. Synthesizing a good grasp proposal for a specific gripper and an object or a scene made of objects is a high-dimensional search or optimization problem as there are many relative gripper-object poses, joint configurations, and contact conditions. The quality of each of these {\em grasp hypotheses\/} can be evaluated under a variety of criteria such as grasp stability, which depends on factors such as object or scene geometry, gripper geometry, and kinematics, as well as suitability for a specific manipulation task. Reflecting on more than four decades of research in robotic grasping, we see a change in how grasping is formulated and studied. 

Early work on robotic grasping developed a theoretical framework that forms the basis of analytical approaches to grasping~\citep{asada1979studies}. At the core of this framework are contact models, which are typically based on point contacts that define what components of contact forces and torques (i.e. wrenches) can be transmitted at a specific contact and act on the object. In this framework, a \textit{grasp} is defined as the set of wrenches that can be achieved on an object. The goal of grasp synthesis is then often framed~\citep{Bicchi2000} as finding a grasp that keeps the object in equilibrium in the presence of disturbances (i.e. {\em fixturing\/}) or moves it in a specific way (i.e. {\em dexterous manipulation\/}). \citet{Bicchi2000} also mention enveloping grasps that wrap the fingers and the palm around the object, achieving more restraining grasps (i.e. {\em power grasps\/}). However, the limitation of these analytical approaches is that they assume full knowledge of object shape and geometry, material properties, and dynamics parameters. But in reality, this information is rarely directly observable but can only be inferred from partial, noisy sensory data.


\begin{figure}[t]
\centering
\includegraphics[width=0.55\linewidth]{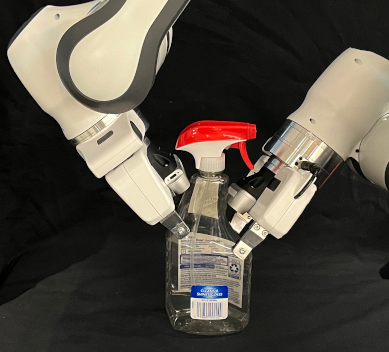}
\caption{Grasp Synthesis is the problem of creating a grasp pose, or a set of grasp poses. Robotic manipulators can grasp objects from multiple angles. This systematic survey reviews deep learning approaches to grasp synthesis that generate grasps utilizing all 6DoF.}
\end{figure}

The increased application of data-driven approaches to computer vision has successfully transferred to robotic grasping~\cite{Bohg_2014}, primarily addressing the complexity and uncertainty in visual perception. To better fit these computer vision techniques, the focus of robotic grasping shifted from concepts around multi-fingered, contact-based representation to pose-based ones. Commonly abstracting the robot's end-effector as an ideal two-fingered gripper approaching the object ``top-down". That way, a grasp is parameterized by the position and orientation of the coordinate frame attached to the gripper or the robot wrist. Before that, the degrees of freedom of a grasp were attributed to the robot hand, its kinematic structure, and the ability to control finger movements. This simplification of grasp parameterization is further supported by the increased availability of robust and simple end-effectors - e.g.\ parallel jaw grippers, suction cups~\cite{FourLessons_2017} - and under-actuated or soft hands~\cite{deimel16-IJRR,Catalano2016} that simplify the control, compared to a dexterous hand. 


\begin{figure}[t!]
\centering
\begin{tabular}{m{1em}cc}
\rowcolors{1}{White}{White}
& Top-Down & 6-DoF Grasp \\
\rotatebox{90}{Dexterous Hand} & \includegraphics[width=.2\linewidth,valign=m, trim=18cm 0.5cm 15cm 18cm, clip]{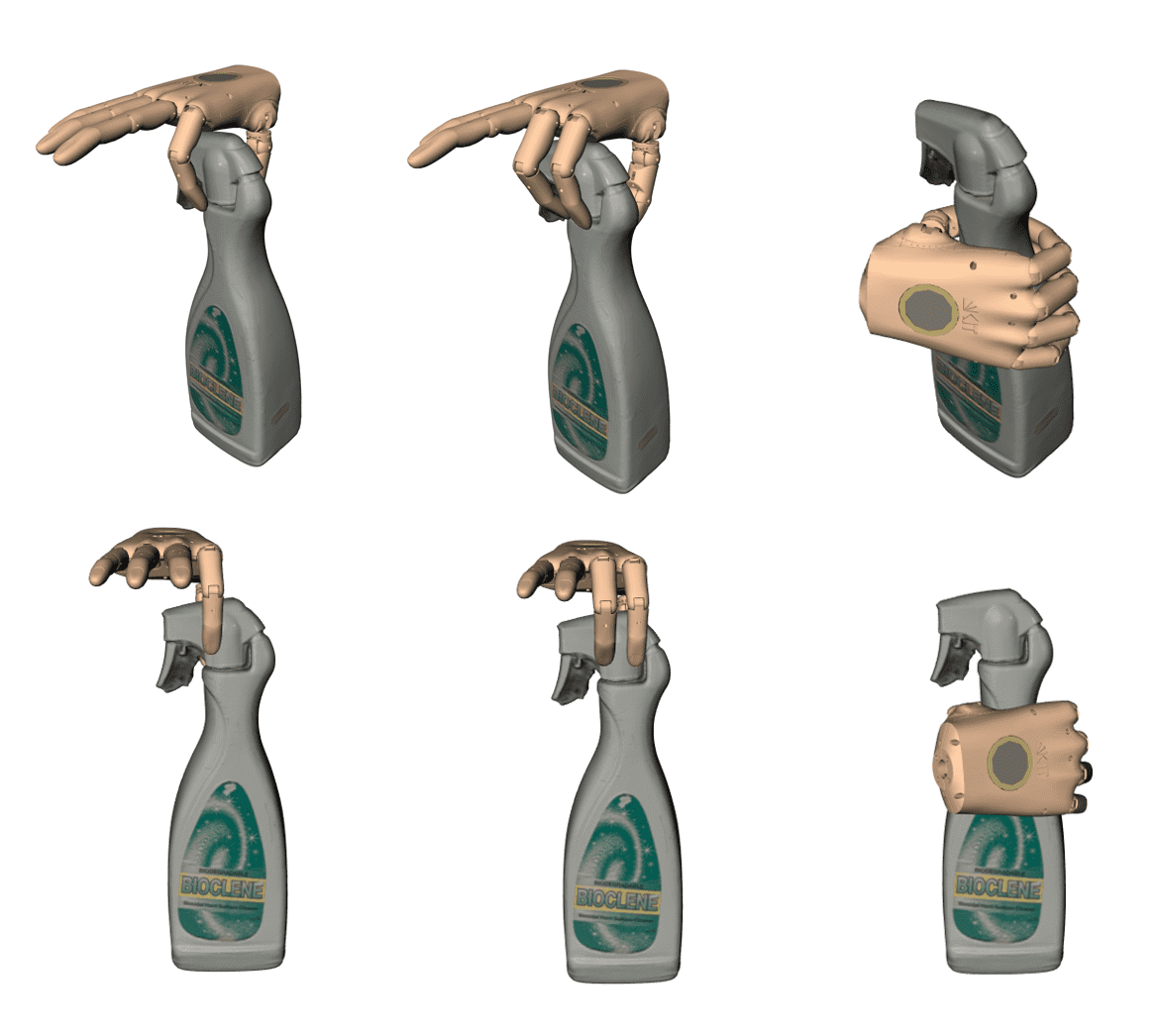} &  \includegraphics[width=.21\linewidth,valign=m, trim=32cm 1cm 1cm 18cm, clip]{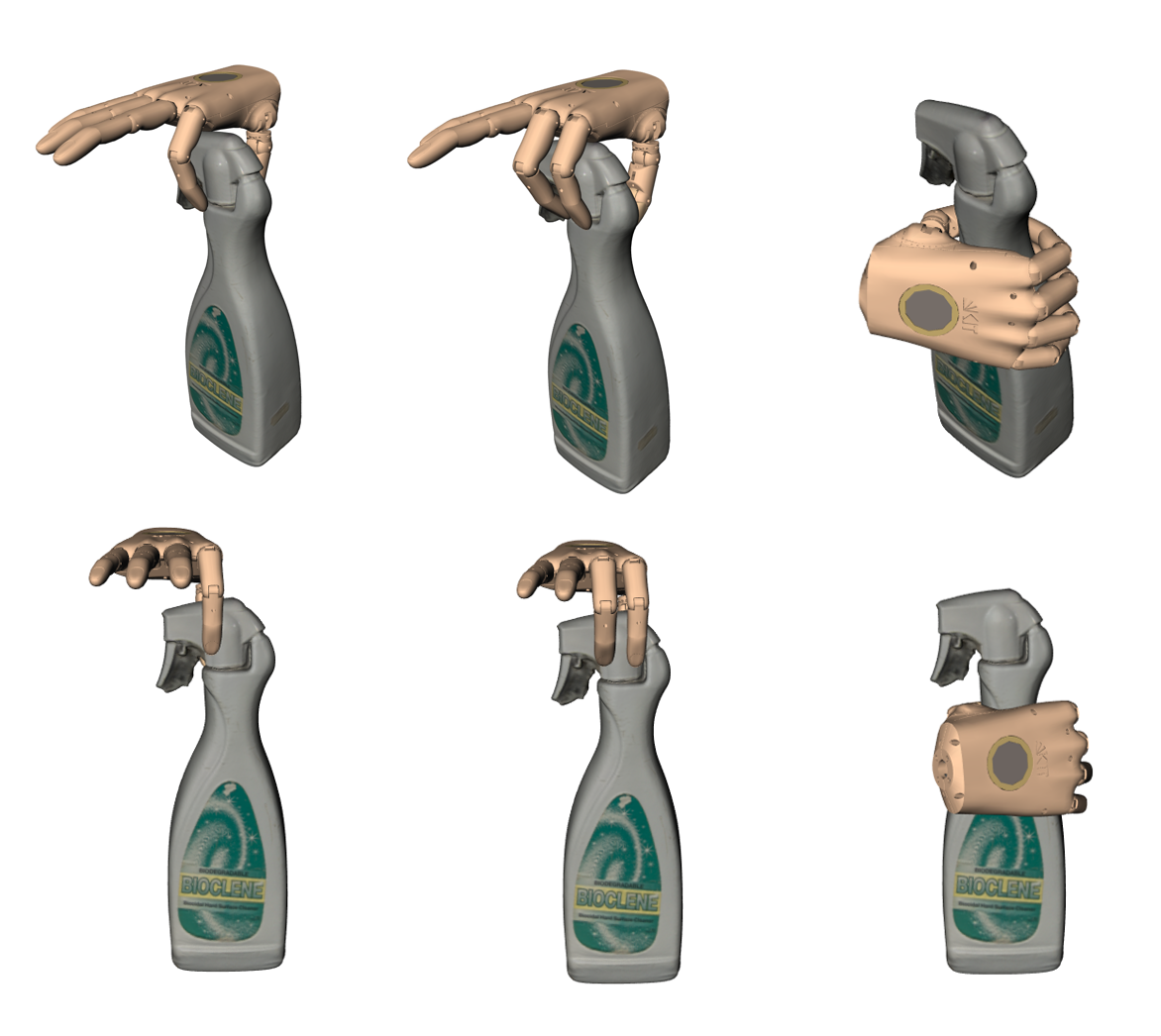}
\global\rownum=0\relax \\
\rotatebox{90}{Parallel Hand} & \includegraphics[width=.4\linewidth,valign=m]{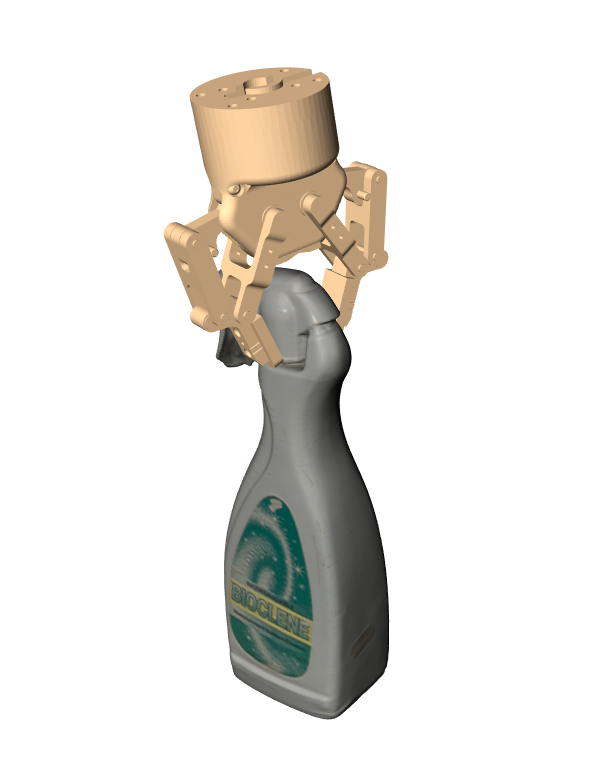} & \includegraphics[width=.4\linewidth,valign=m]{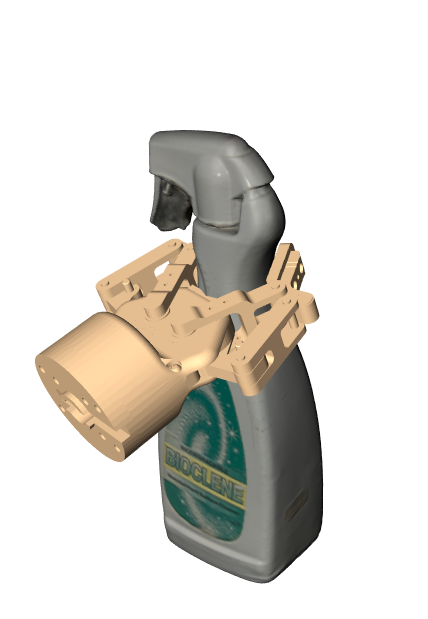}\\
\end{tabular}
\caption{This review focuses on the synthesis of 6 degrees-of-freedom (DoF) grasp pose hypotheses, where the DoF refers to the generated hand/wrist poses defined by the 3D position and orientation of a gripper specific coordinate system. This includes grasp synthesis with parallel grippers and dexterous hands alike.}
\label{fig:grasping_6dof}
\end{figure}

\begin{figure*}[!t]
\centering
\includegraphics[width=1\linewidth]{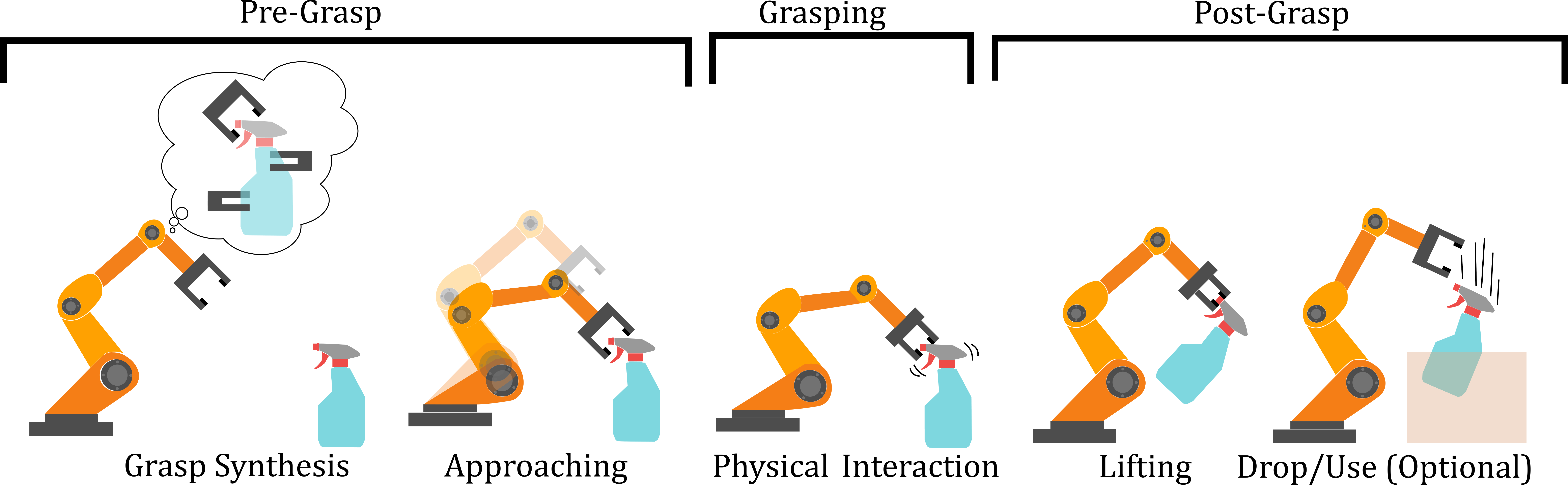}
\caption{Typical stages for grasping an object. Our review focuses on grasp synthesis, the first stage in the grasping process.}
\label{fig:grasping_process}
\end{figure*}


For the purpose of this review, we reuse terms such as four and six degrees of freedom (DoF) grasps, where the former encodes position in 3D and single rotation of a coordinate system attached to the gripper, relevant for defining ``top-down" grasps. Six degrees of freedom grasps relate to approaches that consider the full pose of that coordinate system. \cref{fig:grasping_6dof} shows examples of both 6-DoF and 4-DoF grasps, for both parallel jaw grippers and dexterous hands. While 6 DoF grasp synthesis is particularly useful in applications that require precise and dexterous manipulation of objects, such as for pick-and-place tasks. However, it can also be more challenging than 4-DoF grasp synthesis due to due to the increased complexity of the grasping problem.


The dominant framework in the field of computer vision is applying deep learning to images. The robotics community often benefits from advancements in computer vision as the robots are often equipped with RGB and/or depth cameras. Deep learning, particularly when combined with large-scale data, has enabled new robotic capabilities that traditional machine learning or analytical methods cannot match. These include end-to-end manipulation learning~\citep{levine2016end}, ambidextrous grasping~\citep{mahler2019learning}, in-hand dexterous manipulation~\citep{andrychowicz2020learning}), pick-and-place in clutter~\citep{zeng2022robotic}, as well as more recent works such as task-agnostic generalization to new tasks~\citep{rt12022arxiv} and translating natural language commands to robot actions~\citep{lynch2022interactive}.

Deep learning methods are also crucial for 6-DoF grasp synthesis because they can offer increased accuracy, flexibility, reduced manual engineering, and robustness to variability compared to traditional machine learning or analytical methods. They can learn complex and flexible models that adapt to a wide range of grasping scenarios, without requiring manual feature engineering. Additionally, deep learning methods can account for variations in object shape, pose, and other factors, resulting in more robust grasping performance. Typical stages for grasping an object are illustrated in \cref{fig:grasping_process}, with grasp synthesis being the first stage before any action is taken. In this review, we focus on deep-learning approaches applied to 6-DoF grasping, in particular on the grasp synthesis stage of the grasping process.


There are recently published review papers \cite{Kleeberger2020ASO,Kroemer2021ARO,PointCloudSurvey} similar to ours. \citet{Kleeberger2020ASO} covers a broad range of methods for grasping, including model-based, analytical, optimization-based, and learning-based methods. While both papers cover similar topics, our paper aims to provide a more detailed and comprehensive review on the state-of-the-art for 6-DoF grasping. \citet{Kroemer2021ARO} provides a general review of machine learning techniques for robot manipulation, including grasping, reaching, and other manipulation tasks. The paper provides a high-level view of the field and is less focused on specific grasping methods. Differently from these two papers, ours has a specialized focus on the specific topic of deep learning for grasp synthesis while offering a comprehensive discussion around future research directions on the topic. To our knowledge, \citet{PointCloudSurvey} is the closest to our work in terms of taxonomy, focusing on point cloud-based inputs and depth of discussion. The discussion in \citet{PointCloudSurvey} is mainly around sampling-based methods, which was the dominant method before 2020, their cut-off year for the reviewed papers. In our paper, in addition to Sampling methods (Section~\ref{sec:sampling}), we capture the increasing popularity of end-to-end methods (called Direct Regression in our review - see Section~\ref{subsec:dr}). We also identify supporting methods, such as Affordances (Section~\ref{sec:semantics}) and Shape Approximation (Section~\ref{sec:shape}), that are commonly used together with main grasp synthesis methods. Differently from these review papers, ours provide an analysis on the benchmarking (Section~\ref{sec:benchmarking}) and datasets (Section~\ref{sec:training}) for grasp synthesis. To our knowledge, our review is the only one that focus on 6-DoF grasp synthesis. We believe it is a crucial area of research for future progress in robotic manipulation.

Our systematic review is based on $85$ research publications that employ deep-learning methods for grasp synthesis, clustering the work along common methodologies, data sets, and object-sets used. From the methods' viewpoint, we devise a taxonomy along: Grasp Sampling methods, Direct Regression methods, methods incorporating Reinforcement Learning (RL), and Exemplar methods. Furthermore, we identify methods commonly employed in grasping, such as Shape Completion and Affordances. \cref{fig:mindmap} depicts a visual representation of the structure of the survey paper. 

The contributions of this survey are:

\begin{itemize}
    \item A systematic review of 85 papers, focusing on deep-learning based 6-DoF grasping.
    \item The synthesis of the papers into 10 key takeaways (discussed in \cref{sec:future}) which we consider crucial for future research in robotics and manipulation.
\end{itemize}


\section{Notations and analytical grasping}
To provide a nuanced discussion regarding the contributions of deep-learning based approaches to grasp synthesis, we briefly overview the necessary fundamentals of grasping and review notations and definitions used to define it.



\begin{itemize}

\item A \textbf{Grasp Pose} defines the position and orientation of a grasp. Our survey found many different formulations of grasping but most aim to learn the final pose of the robot to generate a successful grasp.

\item A \textbf{4-DoF grasp} defines a grasp where hand poses are generated and defined by a 3D position and hand orientation about an approach vector that is commonly aligned with the direction of gravity and is therefore often referred to as ``top-down grasping". It is often denoted by ${x,y,z,yaw}$.

\item A \textbf{6-DoF grasp} defines a grasp where hand poses are generated and defined by a 3D position and orientation, thus 6-DOF in total. The major difference to 4DoF grasps is a non-fixed approach vector, providing extra flexibility but increased complexity.

\item \textbf{Affordances} refers to the different tasks which can be achieved with an object~\cite{affordances_pscyh}. This definition has been adopted in previous grasping works (e.g \cite{Ardon2019}).
\end{itemize}

An alternative way to frame a grasp, introduced by \cite{morales2006}, is using the following three terms:

\begin{itemize}
\item An \textbf{approach vector} defines the line along which the gripper or robotic hand approach the target object.

\item The \textbf{grasp center point} is the point in space somewhere along the approach vector where the coordinate frame fixed to the gripper must be positioned before starting to close the fingers.

\item The \textbf{hand orientation} defines how the robot hand is oriented around the approach vector when placed on the grasp center point. 

\end{itemize}

Some important terms used throughout the paper while describing approaches to grasping. 

\begin{itemize}

\item Each point force applied at a contact point on the object surface also generates a torque on the object. A \textbf{wrench} summarizes this pair of force and torque applied to the object through a contact in a six-dimensional vector. 


\item A grasp is said to be in \textbf{Force Closure} if the forces that can be applied at the set of {\em frictional\/} contacts are sufficient to compensate for {\em any\/} external wrench applied to the object \cite{Nguyen1987}.




\item \textbf{Hand posture} describes the configuration of the gripper or hand fingers when the grasp is started or all the contacts are made

\item A \textbf{Power Grasp} is a
grasp where there are multiple points of contact between the object and the fingers and palm. It maximizes the load carrying ability of a grasp and is highly stable as the enveloping nature of the grasp provides form closure~\cite{powergrasp}.

\item \textbf{Antipodal points} are pairs of points on the object surface whose normal vectors are collinear and pointing in the opposite direction~\cite{Chen1993}. With appropriate finger contact conditions, antipodal point grasps guarantee force closure. 

\item An \textbf{Antipodal grasp} is defined for two-fingered grippers that makes contact with the object at antipodal points~\cite{Chen1993}.

\end{itemize}

For consistency and a thorough discussion, we provide a short insight on the relevance of analytical approaches.

The majority of methods up to the year 2000 modeled grasping analytically~\cite{Bicchi2000, Bohg_2014}. The focus was on modeling and estimating physical conditions of grasps, such as, for example, grasp stability. A force-closure grasp was often equated with a stable grasp, although force-closure is a necessary but insufficient condition for a stable grasp~\cite{Bicchi2000, prattichizzo2016grasping}. 
Physical conditions were usually simplified through approximations such as point contact models, Coulomb friction, and rigid body dynamics~\cite{Murray1994AMI, Bohg_2014, prattichizzo2016grasping}. 
Analytic approaches have been attributed to being complex and not applicable in real-time applications. However, analytic approaches address properties of grasps, while most of the recent methods for grasp synthesis focus on positioning the hand. The advantages of analytical methods are mathematical guarantees on grasp properties, such as force-closure. This makes it easier to assess the conditions of grasps when objects are manipulated after a grasp has been applied. For example, what forces or torques can be exerted on the object before slippage occurs or how an object can be moved using in-hand manipulation. Within the first decade of the 21st century, there has been a rise of data-driven approaches to grasp synthesis~\cite{Bohg_2014}, thanks to the development of grasping simulators such as GraspIt!~\cite{GraspIt}. Early approaches often used hand-designed features that corresponded to parts of objects that could be grasped~\cite{elkhoury2007, Kyota2005, cedric2006}. \citet{Kamon1996} presented one of the earliest works in 4-DoF grasping that use machine learning (ML) for grasping objects. The authors hand designed a low-dimensional feature space to estimate the quality of grasps. Since then, many works have employed traditional learning methods with a larger feature space~\cite{Jiang2011, Saxena2006LearningTG, Saxena2008, Saxena2008LearningGS, Morales2004}. One key takeaway from this early research was that while grasping can be done using RGB data only~\cite{Saxena2008}, depth information (RGB-D) improved grasping success~\cite{Jiang2011, Saxena2008LearningGS}. We refer to \cite{Bohg_2014} for a more in-depth review of earlier methods for data-driven grasp synthesis. In this review, we focus on deep learning techniques in particular.

\begin{figure}
    \centering
    \pgfplotstableread[col sep=comma]{images/growth.csv}\data

\begin{tikzpicture}
    
    \begin{axis}[
        name=mainplot,
        xmin = 2000, xmax = 2021,
        ymin = 0, ymax = 120,
        enlarge y limits=true,
        width = 0.5\textwidth,
        ymajorgrids,
        height = 0.28\textwidth,                
        ylabel= \# Papers,
        xtick=data,
        xtick style={draw=none},
        xticklabel style={/pgf/number format/1000 sep=,rotate=60,anchor=east,font=\scriptsize},
        y label style={at={(axis description cs:-0.075,.5)},anchor=south},
        ]
        
        \addplot[red, mark=none] table[x=y , y=n] {\data};
        \addplot[mark=none, draw=none, name path=A] coordinates {(2012,-200) (2012,200)};
        \addplot[mark=none, draw=none, name path=B] coordinates {(2021,-200) (2021,200)};
        \addplot+[green, fill opacity=0.2] fill between[of=A and B,soft clip={domain=1990:2030}];
        \node[scale=0.9] at (axis cs: 2016.5,110) {{\small Span Of Our Review}};
        \end{axis}
\end{tikzpicture}
    \caption{The number of publications on IEEExplore that includes the keyword ``Grasping" in metadata and ``6DoF" in the full text is increasing year-by-year. We consider works published after Jan 1, 2012 -- when AlexNet~\cite{krizhevsky2012imagenet} was published -- in our review.}
    \label{fig:publication_growth}
\end{figure}
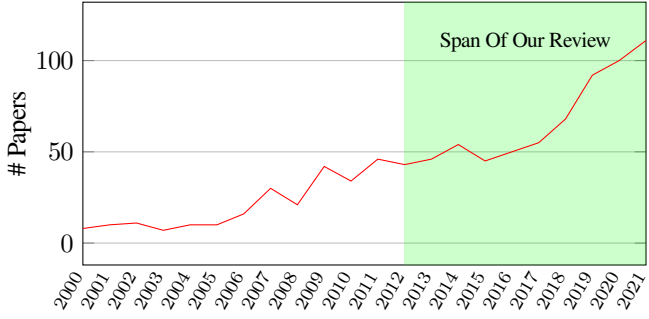 
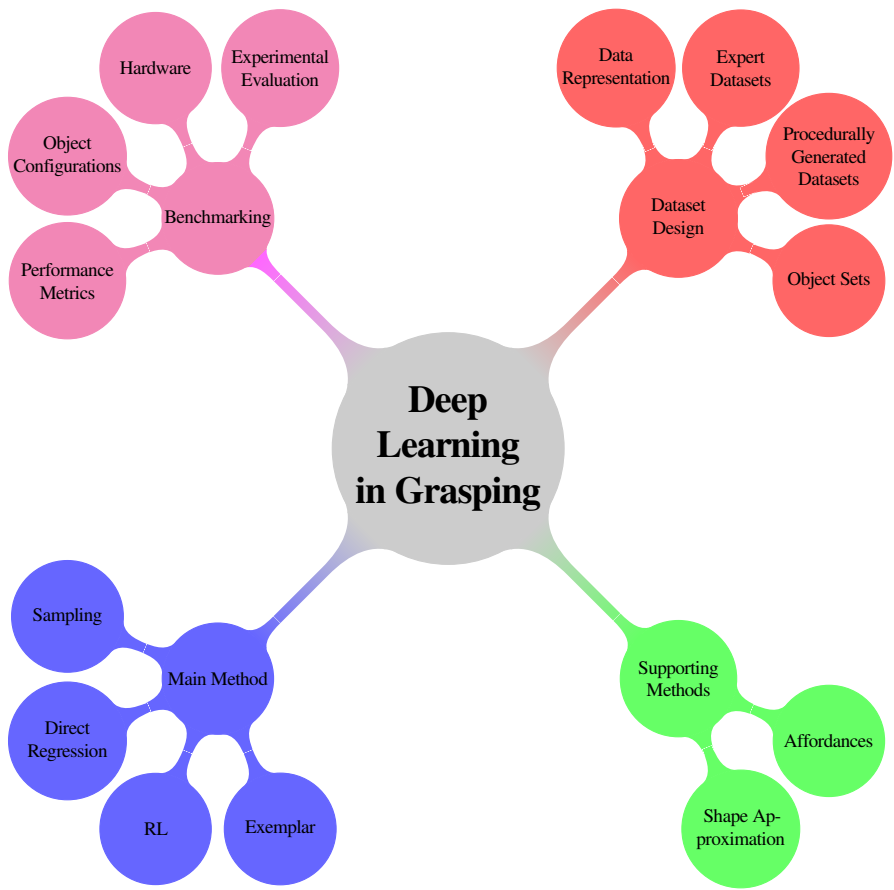
\begin{figure*}[t!bh]
\centering
\resizebox{0.65\textwidth}{!}{
\begin{tikzpicture}[
    mindmap,
    grow cyclic, text width=4cm, align=flush center,
    every node/.style={concept},
    font=\huge,
    concept color=gray!40,
    level 1/.style={level distance=12cm,sibling angle=90},
    level 2/.style={level distance=6cm,sibling angle=45},
    level 3/.style={level distance=6cm,sibling angle=45}]]

\node [root concept, scale=2, font=\huge] {\textbf{Deep Learning in Grasping}}
   child [concept color=blue!60] { node {Main Method}
        child { node {Sampling}}
        child { node {Direct Regression}}
        child { node {RL}}
        child { node {Exemplar}}
    }
   child [concept color=green!60] { node {Supporting Methods}
       child { node {Shape Approximation}}
       child { node {Affordances}}
    }
   child [concept color=red!60] { node {Dataset Design}
      child {node {Object Sets }}
      child {node {Procedurally Generated Datasets}}
      child {node {Expert Datasets}}
      child {node {Data Representation}}
    }   
    child [concept color=magenta!60] { node {Benchmarking}
    child { node {Experimental Evaluation}}
    child { node {Hardware}}
    child { node {Object Configurations}}
    child { node {Performance Metrics}}
    }
    ;
\end{tikzpicture}}
\caption{A visual representation of the core topics covered in this systematic survey.}
\label{fig:mindmap}
\end{figure*}

\section{Deep Learning Methods in 6-DoF Grasping}
\label{sec:deep}

The number of publications investigating deep learning approaches for 6-DoF grasping have grown significantly in the last few years, as highlighted in \cref{fig:publication_growth}. From the systematic review, we identify four main algorithmic methodologies for grasp synthesis using deep-learning based on the reviewed literature: grasp pose sampling, regressing grasp pose directly, reinforcement learning and exemplar methods. These procedures relate to how the grasps are generated at test time.   

 
Sampling approaches consider one or many grasp samples and have learned a function to estimate the quality of a sampled grasp. An essential characteristic of sampling approaches is that each sample is evaluated \textit{individually}. Alternatively, direct regression considers the data \textit{globally} and learns a function to predict high-quality grasps. \textit{RL} includes methods that involve the maximization of a cumulative reward function based on a robot's actions or demonstrations. Exemplar methods aim to use a similarity metric between grasps to retrieve high-quality grasps from an existing database which are most similar. These categories are often mutually exclusive, though \citet{aktas2019deep} employed both direct regression and sampling while \citet{Dexnet1} employed both sampling and exemplar methods.

\subsection{Sampling}
\label{sec:sampling}
\begin{table}[htb]
\centering
\caption{Publications with deep-learning focused sampling methods. We cluster the papers based on the space the sample through and how the samples are evaluated. Some approaches further consider an optional refinement stage.}
\begin{tabular} {| c | c ? c | c | c |}
\hline
\rowcolor[HTML]{D0CECE}

Year & Paper & Sample Space & Evaluation & Refinement \\ \hline
2015 & \citet{varley2015generating} & Priors & Metric & \\ \hline
2015 & \citet{bohg2015dataset} & \textit{Not Specified} & Binary & \\ \hline
2016 & \citet{gualtieri2017high} & Euclidian & Binary & \\ \hline
2017 & \citet{ten2017grasp} & Euclidian & Binary & \\ \hline
2017 & \citet{zhou20176dof} & Euclidian & Binary & \ForestGreencheck\\ \hline
2018 & \citet{yan2018learning} & \textit{Not Specified} & Binary & \ForestGreencheck\\ \hline
2019 & \citet{mousavian20196} & Latent & Binary & \ForestGreencheck\\ \hline
2019 & \citet{Liang_2019} & Priors & Metric & \\ \hline
2019 & \citet{lu2019planning} & Euclidian & Binary & \ForestGreencheck\\ \hline
2019 & \citet{ottenhaus2019visuo} & Priors & Metric & \\ \hline
2019 & \citet{Goncalves2019} & Euclidian & Binary & \\ \hline
2019 & \citet{aktas2019deep} & Priors & Binary & \ForestGreencheck\\ \hline
2020 & \citet{Riedlinger2020} & Euclidian & Binary & \\ \hline
2020 & \citet{murali20206} & Latent & Metric & \ForestGreencheck \\ \hline
2020 & \citet{vandermerwe2020learning} & Hand Posture & Binary & \ForestGreencheck\\ \hline
2020 & \citet{lundell2020beyond} & Multiple Views & Binary & \\ \hline
2020 & \citet{lou2020learning} & Euclidian & Metric & \\ \hline
2020 & \citet{choi2020hierarchical} & Multiple Views & Binary & \\ \hline
2020 & \citet{lu2020multifingered} & Hand Posture & Binary & \ForestGreencheck\\ \hline
2020 & \citet{Schaub2020} & Multiple Views & Binary & \\ \hline
2020 & \citet{murali2020same} & Priors & Metric & \\ \hline
2020 & \citet{Kokic2020} & Euclidian & Metric & \\ \hline
2021 & \citet{lundell2021multifingan} & Euclidian & Metric & \\ \hline
2021 & \citet{lou2021collisionaware} & Euclidian & Metric & \\ \hline
2021 & \citet{lundell2021ddgc} & Euclidian & Metric & \\ \hline
2021 & \citet{peng2021selfsupervised} & \textit{Not Specified} & Metric & \\ \hline
2021 & \citet{jiang2021synergies} & Euclidian & Binary & \\ \hline
2021 & \citet{kasaei2021mvgrasp} & Multiple Views & Binary & \\ \hline
2021 & \citet{Wang2021} & Priors & Binary & \\ \hline
2021 & \citet{munoz2021grasping} & Multiple Views & Binary & \\ \hline
2021 & \citet{corsaro2021learning} & Priors & Binary & \\ \hline
2021 & \citet{Ren2021} & \textit{Not Specified} & Metric & \\ \hline
2021 & \citet{wencatgrasp} & \textit{Not Specified} & Metric & \\ \hline

\end{tabular}
\label{tab:SamplingPapers}
\end{table}


We define sampling methods as any approach that considers each sample \textit{individually} and use information encoded about the sample to make decisions about the grasp. The samples may be sourced from any discrete or continuous $n$-dimensional space. All reviewed sampling works are shown in \cref{tab:SamplingPapers}.


Deep-learning approaches employing sampling implement the following steps: sample information; evaluate the sample according to a quality estimation function, parameterized by a deep neural network; and optionally refine the sample using an optimization-based approach to achieve a higher quality grasp. We adopt the term `quality', however, there is no current consensus on the definition. Throughout this paper, we use this term to refer to the confidence of grasp success. Table~\ref{tab:SamplingPapers} presents an overview of all the \textit{Sampling} papers. A popular deep-learning based approach for 4-DOF grasping generates a series of antipodal grasps through sampling while using a neural network to predict grasp quality~\cite{mahler2018dexnet}. This aims to predict the probability of a successful grasp to generate a series of antipodal grasps.



\subsubsection{Generating Samples}

Authors can sample a subset of grasp parameters in an n-dimensional space. This is commonly done using one of two approaches. Random sampling occurs when samples are taken from an arbitrary random distribution. The most common distribution is a uniform distribution, however, some authors sample from other distributions such as Gaussian~\cite{vandermerwe2020learning, lu2020multi}. The second method is where samples are taken at equispaced intervals within the space.

The subset of grasp parameters can be sampled from: Euclidean space, priors, configuration space, latent space, or multiple views. When sampling in Euclidean space, heuristic-based rules are often used to remove irrelevant grasp candidates. 

\paragraph{Euclidean Space Sampling}
One of the most common approaches is to sample a 3D vector representing the translation part of the grasp pose. This is often achieved by sampling points from a point cloud~\cite{gualtieri2017high, lou2020learning, lou2021collisionaware}. The approach vector can be estimated by either using normal information estimated from the sensor data or sampling angles. 
\citet{ten2017grasp} and \citet{Riedlinger2020} use the opposite direction of the surface normal to find the grasp approach vector. \citet{Riedlinger2020} uses a series of local augmentations on the initial grasp approach vector to generate a sample set of candidate approach vectors. \citet{gualtieri2017high} generate candidate grasp approach vectors using the surface normal and curvature axis to generate equispaced orientations orthogonal to the curvature axis. 

Some approaches sample angles independently of the surface normals. \citet{lou2020learning, lou2021collisionaware} chose $N$ points from a point cloud and sample the wrist angles randomly for the grasp. However, they restrict the approaching vector of the grasp to be above the table. \citet{Kokic2020} randomly sample grasp and roll angles, and offset distances for each point in the point cloud. Both \citet{lu2019planning, lundell2021multifingan, lundell2021ddgc} sample grasp candidates around the center of an object with a random orientation. 

Sampling can also be performed with regularly spaced points through euclidean space. \citet{jiang2021synergies} sample regularly spaced points for the position of the grasp pose. Similarly, \citet{Goncalves2019} sample equispaced points throughout the region of interest.

Alternatively, instead of sampling a $xyz$ vector, angles can be used. \citet{zhou20176dof} chose a random hand orientation and approach vector after which they translate the hand along the approach vector of each sample until a grasp is found that does not collide with the gripper when it is open but collides when closed. 

In Euclidean space, sampled grasp are commonly pruned to remove infeasible grasps based on rules such as: 

\begin{itemize}
    \item The robot hand is in collision with the point cloud when fingers are open~\cite{ten2017grasp, zhou20176dof}.
    \item The closing region of the fingers does not contain at least one point from the point cloud~\cite{ten2017grasp, gualtieri2017high, zhou20176dof}. 
\end{itemize}

\paragraph{Sampling Priors}

More complex algorithms can also be used to find a set of feasible grasp candidates. These are often used when creating grasp samples which will then be evaluated for affordances~\cite{murali2020same} or different types of grasps (e.g power, pinch grasps)~\cite{corsaro2021learning}. The most commonly~\cite{murali2020same, Liang_2019, Wang2021, corsaro2021learning} used grasping algorithm is an SVM-based approach proposed by \citet{pas2015using}. \citet{Dexnet1} use a modification of the grasping algorithm proposed by \citet{Smith1999ComputingPG} to generate a series of antipodal grasps. They frame the problem as a multi-armed bandit, to be solved using Thompson sampling~\cite{agrawal2012analysis}.

While grasping simulators are often used for generating training data, they have also been used at test time to sample grasps. A common prerequisite for grasping simulators is a full approximation of the shape model. The common simulators for this purpose were GraspIt!~\cite{GraspIt} or Simox~\cite{Vahrenkamp2013}. \citet{ottenhaus2019visuo} generated grasp samples on a reconstructed object using the grasp planner by Simox~\cite{Vahrenkamp2013}. Alternatively, \citet{varley2015generating} sampled grasps using the Simulated Annealing planner~\cite{Ciocarlie2009} for partially visible objects.

A deep-learning approach can also be used to synthesize grasp samples. \citet{aktas2019deep} used the direct regression based approach from \citet{kopicki_dexterous}  to generate multiple grasps, to be used as grasp samples.

\paragraph{Latent Space Sampling}

\begin{figure}
    \centering
    \includegraphics[width=0.75\linewidth]{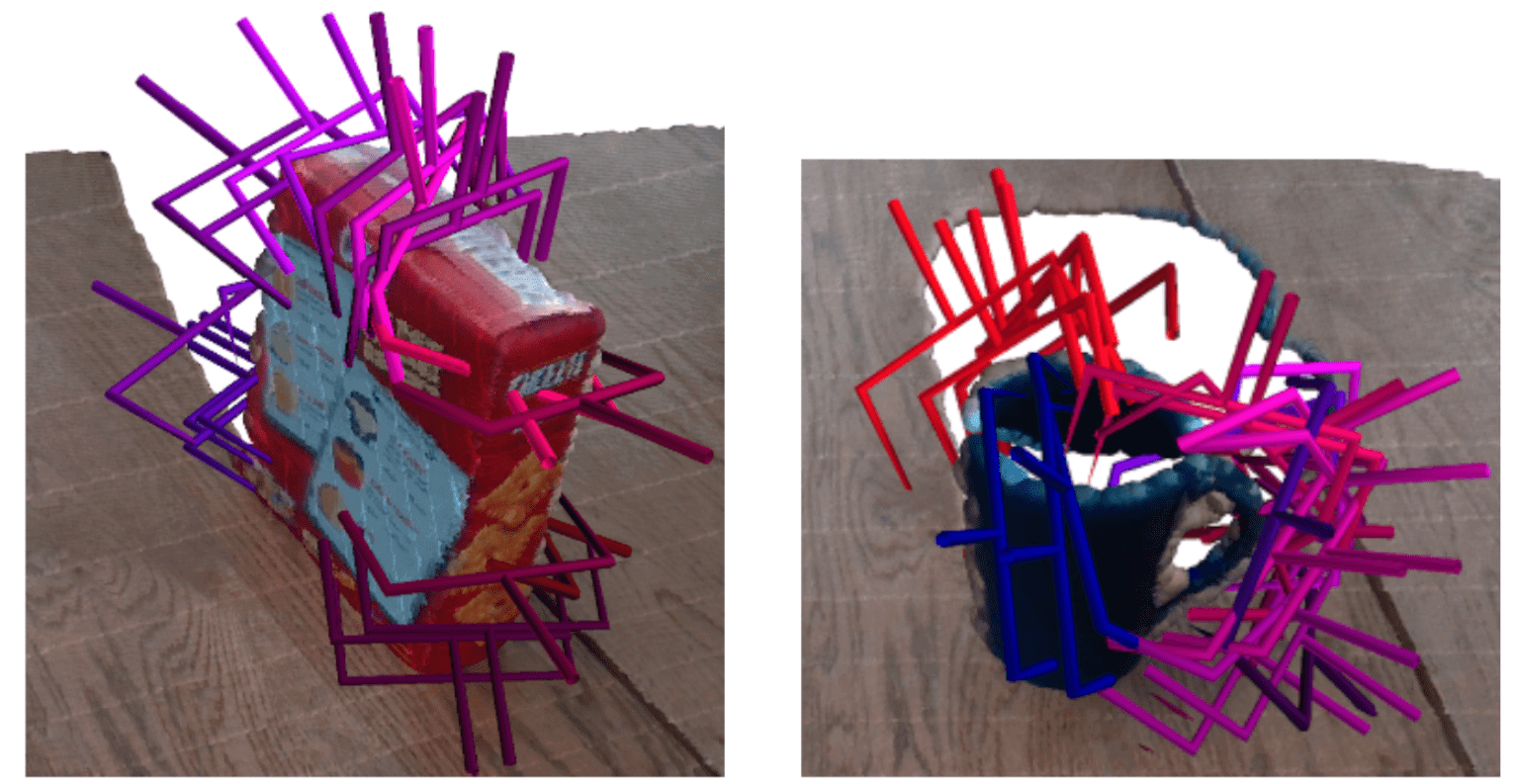}
    \caption{\citet{mousavian20196} sample through the latent space of a trained model to generate a series of grasp candidates ($\copyright$2019 IEEE)}
    \label{fig:latent_space}
\end{figure}

\citet{mousavian20196} train a Variational Auto Encoder~\cite{kingma2014autoencoding} and uniformly sample through latent space to generate grasp poses. An example of the grasps generated by this approach is shown in \cref{fig:latent_space}. This sampling approach is also adopted by \citet{murali20206}.  

\paragraph{Hand Posture Space Sampling}

Some authors sample grasp configurations from a prior distribution fit to the training set to create an initial hand configuration for their approach~\cite{vandermerwe2020learning, lu2020multifingered, lu2020multi}. \citet{vandermerwe2020learning} and \cite{lu2020multi} fitted a Gaussian Mixture Model to represent a grasp prior function trained on grasp configurations seen during training. Similarly, \citet{lu2020multifingered} trained a Mixture Density Network~\cite{bishop1994mixture} over all grasp attempts from the training set. 

\paragraph{Multiple Views}

A number of papers employ multiple viewpoints of the scene, where these can either be from virtual or real cameras. A camera viewpoint is then sampled, and a grasping approach is employed on the sampled viewpoint. \citet{Schaub2020} combine multiple viewpoints around the scene to create a 3D representation of an object. Using the 3D representation, they generate depth images for a series of virtual cameras. They then use a 4-DoF grasping algorithm~\cite{morrison2018closing} for each real and virtual camera. This is extended by \citet{Schuab2021} who fused depth images around the scene to provide more detailed depth maps. \citet{choi2020hierarchical} extended a previous 4-DoF approach~\cite{Satish2019} to 6-DoF by using an iterative improvement algorithm approach to choose an approach direction. \citet{munoz2021grasping} and \citet{kasaei2021mvgrasp} both generate multiple views of the object from virtual cameras using a captured point cloud from a single viewpoint. They proposed a method to select a view according to an entropy-based measure. \citet{lundell2020beyond} use the algorithm proposed by \citet{Satish2019} on the depth map from multiple viewpoints. The robot then executes the best grasp from all viewpoints.

\subsubsection{Sample Evaluation} 
\label{sec:Evaluating_Samples}

Once samples are generated, these approaches use a function, commonly parameterized by a neural network, to estimate a numerical value representing a grasp metric. This is commonly used an estimation of the `quality' of the grasp. The grasp quality can be approximated through an analytical metric or the likelihood of a grasp being successful. While some approaches execute the highest quality sample directly, others refine the samples using optimization-based techniques based on the learned quality function. 


\paragraph{Binary classification} Predicting grasp success can be treated as a binary classification problem, where the output represents the confidence of a grasp being successful or not. A CNN can be trained for the binary classification of grasp success, with the network either outputting 0 or 1 for a successful grasp~\cite{ten2017grasp, Riedlinger2020, gualtieri2017high, Goncalves2019, corsaro2021learning,bohg2015dataset, mousavian20196, vandermerwe2020learning, lu2019planning, zhou20176dof, aktas2019deep, lu2020multifingered, yan2018learning, Wang2021}. During the collection of ground-truth training data to determine if a grasp is successful, the grasp is most commonly executed using a robotics simulation environment~\cite{jiang2021synergies, Riedlinger2020, corsaro2021learning, mousavian20196, bohg2015dataset, vandermerwe2020learning, lu2019planning, zhou20176dof, aktas2019deep, lu2020multifingered, yan2018learning}, where a grasp will be labeled successful if the object remains within the gripper after being lifted. Grasp success can also be based on an analytical metric, such as whether the grasp would form a force-closure grasp~\cite{ten2017grasp, Goncalves2019} or if the grasp is antipodal~\cite{gualtieri2017high}.

The binary classification of a sampled grasps can also be consider as one of the outputs of a sampling approach. For example. \citet{jiang2021synergies} designed an approach which learns implicit functions that predicts grasp parameters (quality, rotation and width) from the feature space representation of a randomly sampled query point. \citet{vandermerwe2020learning} trained a network to predict the success of a grasp, as well as a signed distance function that represents a distance between a query point and the surface of the object.

\paragraph{Learning a metric} Learning metrics associated with grasps, rather than a pure binary label, have been proposed to better represent the quality of a grasp. These metrics are often continuous numbers (rather then the previously described binary label) that provide additional information to rank grasp quality or help ``guide'' deep learning algorithms using them as a fitness score.

\citet{varley2015generating} learn a series of heatmaps, where each pixel represents the location's efficacy as a fingertip or palm location for common grasp types found in their training set. \citet{Liang_2019} designed an approach which learns a grasp quality metric based on the force-closure metric and wrench space analysis~\cite{Ferrari1992}. Similarly, \citet{ottenhaus2019visuo} train a CNN to estimate the force-closure probability of a grasp under a small random perturbation. \citet{lundell2021multifingan} and \citet{lundell2021ddgc} trained a grasp classifier using a {\em Generative Adversarial Networks\/} (GANs), and use the discriminator loss to help produce realistic-looking grasps. \citet{wencatgrasp} compute a continuous score for each grasp by looking at the stability of randomly sampled grasps in the proximity of the selected grasp. They argue that ``grasp stability should be continuous over its 6D neighborhood", therefore, this should allow for more stable grasps. Furthermore, this can allow for imperfections in the robotic grasp position whilst still being likely to execute a stable grasp. \citet{peng2021selfsupervised} and \citet{Ren2021} designed an approach to learn the smallest co-efficient of friction which will satisfy a force-closure grasp. Learned metrics can also represent the quality of the grasp with respect to other aspects of the task, for example, metrics based on the relevance of the grasp to a given affordance~\cite{murali2020same, Kokic2020}.

Other approaches consider workspace constraints. \citet{lou2020learning} designed an approach to learn the probability of grasp success and use an additional network to learn the probability of a grasp being reachable. These probabilities can be multiplied to find the likelihood of success for the entire grasping action. This work is extended by \citet{lou2021collisionaware} to allow the robot to grasp in constrained environments, such as in boxes, where walls may limit the success of grasping an object.



\subsubsection{Optimization-based Grasp Refinement} %

\begin{figure}
    \centering
    \includegraphics[width=0.75\linewidth, trim={0 0 0 12cm}, clip]{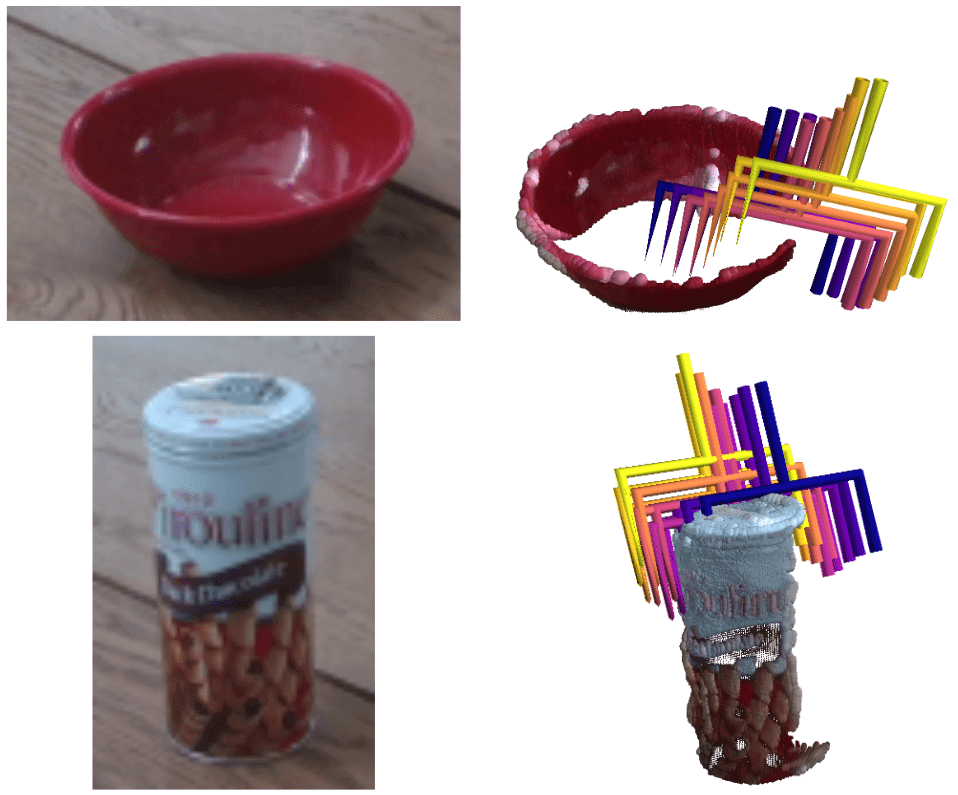}
    \caption{An example of a grasp refinement process from ~\cite{mousavian20196}, where the initial grasp is shown in dark blue and the final grasp pose is shown in yellow. ($\copyright$2019 IEEE)} 
    \label{fig:refine}
\end{figure}

Gradient-based optimization through the trained network can be used to find high-quality grasps candidates. A sampled grasp is taken as the initial seed, which is then refined based on the derivatives of the quality estimation network. This attempts to maximize estimated grasp quality. \citet{zhou20176dof} train a CNN to predict grasp quality given a depth image and a sampled end-effector pose. The end-effector pose is locally optimized using the quasi-newton method on the gradient of the learned quality function. Similarly, some authors take the derivative of the grasp quality with respect to the grasp pose and then use gradient ascent to refine the grasp candidates~\cite{lu2019planning, yan2018learning, mousavian20196, murali20206}. An example of the refinement process is shown in \cref{fig:refine}, where the initial grasp is a potentially bad sample (blue) and is refined to a higher quality grasp (yellow). Both \citet{vandermerwe2020learning} and \citet{lu2020multifingered} treat finding a grasp configuration as an optimization problem aiming to maximize the probability of grasp success, seeding the process with samples from a prior distribution. \citet{lu2020multi} extended this using active learning during training, improving their results. In contrast to other works, \citet{aktas2019deep} found that optimization of the sample did not improve the grasp success rate in simulation. The authors explore the use of gradient-based optimization and simulated annealing for their optimization.




\subsection{Direct Regression}
\label{subsec:dr}

\begin{table}
\centering
\caption{Approaches to direct regression of 6-DoF grasping using Deep Learning, either generating a single or multiple grasp poses as an output. We found three main approaches: regress to a pose directly (Pose), employing a multi-stage approach, or perform dimensionality reduction (DimRed). }
\begin{tabular} {| c | c ? c | c | c | c |}
\hline
\rowcolor[HTML]{D0CECE}

Year & Paper & Grasps & Direct Pose & Multi-stage & DimRed \\ \hline
2017 & \citet{Veres2017} & Single & \ForestGreencheck &  & \\ \hline
2018 & \citet{Schmidt2018} & Single & \ForestGreencheck &  & \\ \hline
2018 & \citet{choi2018soft} & Single &  &  & \ForestGreencheck\\ \hline
2019 & \citet{liu2019generating} & Single & \ForestGreencheck &  & \\ \hline
2020 & \citet{qin2020s4g} & Multiple &  &  & \ForestGreencheck\\ \hline
2020 & \citet{yang2020robotic} & Single & \ForestGreencheck &  & \\ \hline
2020 & \citet{jeng2020gdn} & Multiple &  &  & \ForestGreencheck\\ \hline
2020 & \citet{Wang2020} & Single &  &  & \ForestGreencheck\\ \hline
2020 & \citet{fang2020graspnet} & Multiple &  & \ForestGreencheck & \\ \hline
2020 & \citet{wu2020grasp} & Multiple &  &  & \ForestGreencheck\\ \hline
2020 & \citet{ni2020pointnet} & Multiple &  &  & \ForestGreencheck\\ \hline
2020 & \citet{breyer2021volumetric} & Multiple &  &  & \ForestGreencheck\\ \hline
2020 & \citet{liu2020deep} & Single & \ForestGreencheck &  & \\ \hline
2020 & \citet{shao2020unigrasp} & Single & & \ForestGreencheck & \\ \hline
2020 & \citet{ni2021learning} & Multiple &  & \ForestGreencheck &\\ \hline
2021 & \citet{Wang_2021_ICCV} & Multiple &  & \ForestGreencheck & \\ \hline
2021 & \citet{sundermeyer2021contactgraspnet} & Multiple &  &  & \ForestGreencheck\\ \hline
2021 & \citet{zhao2020regnet} & Multiple &  & \ForestGreencheck & \\ \hline
2021 & \citet{gou2021rgb} & Multiple &  &  & \ForestGreencheck\\ \hline
2021 & \citet{zhu20216dof} & Multiple &  & \ForestGreencheck & \\ \hline
2021 & \citet{wei2021gpr} & Multiple &  & \ForestGreencheck & \\ \hline
2021 & \citet{li2021simultaneous} & Multiple &  &  & \ForestGreencheck\\ \hline
2021 & \citet{Li2021} & Single &  &  \ForestGreencheck &\\ \hline

\end{tabular}

\label{tab:direct_regress}
\end{table}

Direct regression approaches can simultaneously process the entire sample space using a single network to predict either a single or multiple grasps, along with specific grasp properties such as parameters and quality, from visual information.

Direct regression approaches, which can be considered as end-to-end methods, utilize a single network to process the entire input to regress an output. These methods attempt to reduce computational cost compared to sampling methods by processing data globally through the network in a single pass. Early works in direct regression based on deep learning are inspired by object detection work from the Computer Vision community~\cite{yolo, ren2016faster}. Authors would treat finding top-down grasps similar to detecting objects in an image and would use the depth image to recover the 3D position of the grasp~\cite{redmon2015realtime, lenz2014deep, chu2018realworld, kumra2017robotic}. \citet{morrison2018closing} designed a heat-map based approach for 4-DoF grasping that was designed for fast inference, enabling closed-loop grasping and grasping of moving objects. However, as the dimensionality of the output increased, the problem difficulty also increased~\cite{mousavian20196, wu2020generative}. This presents scaling issues when directly regressing to high-dimensional outputs such as a 6-DoF grasp pose or a high-DoF dexterous robotic hand configuration. To overcome this, researchers often reduce the DoF of the output from a single network. This can be accomplished through multi-stage approaches, where each stage has a specific task, or by reducing the DoF by conditioning the output on an input. The direct regression approaches found in our systematic review are shown in \cref{tab:direct_regress}.

\paragraph{Directly Regressing a Pose} \citet{Schmidt2018} presented one of the earliest CNN-based works to directly estimate a single 6-DoF grasp pose of an end-effector from input depth images. Similarly, \citet{yang2020robotic} trained a network to estimate the transformation matrix needed to be applied to the end-effector to produce a successful grasp. However, both approaches assume that there is only a single, most optimal ground truth grasp for each input. This would introduce ambiguities as there may potentially be numerous successful grasps that can be executed for any input. To account for this, \citet{liu2019generating} designed a loss function that accounts for multiple ground truth grasps in the training data by calculating the loss between the current output and the closest ground truth grasp. This is extended in another work~\cite{liu2020deep} that includes differentiable terms for a grasp metric and self-collisions in the loss function. This allows their approach to work both in an unsupervised manner or using a smaller supervised dataset. \citet{Veres2017} create a generative model based on a {\em Conditional Variational Autoencoder\/} (CVAE)~\cite{Sohn2015LearningSO}. They use the CVAE to generate contact positions and contact normals for a multi-fingered robotic hand.

\begin{figure}
    \centering
    \includegraphics[width=1\linewidth, trim={1cm, 0, 28cm, 0}, clip]{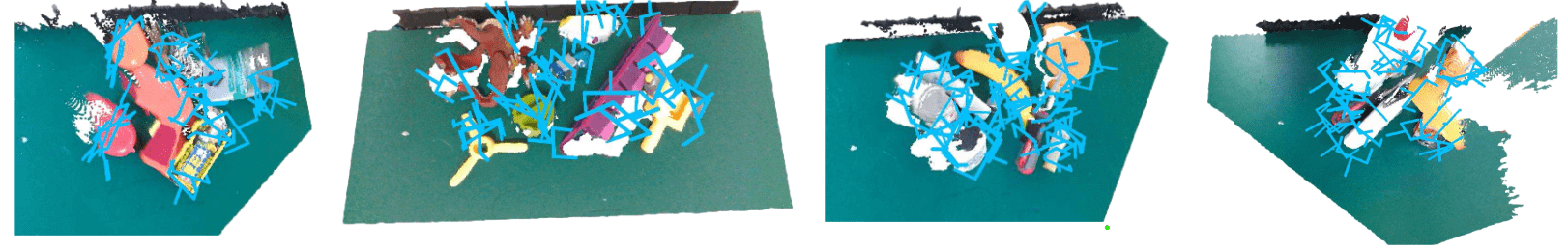}
    \caption{\citet{fang2020graspnet} uses a direct regression approach to generate a series of grasps from real-world data. ($\copyright$2020 IEEE)}
    \label{fig:direct}
\end{figure}

\paragraph{Reducing DoF}

Due to the difficulty of regressing all 6 DoF of a grasp, some DoF can be reduced by analytically determining some DoF conditioned on the regressed DoF. This is seen in some 4-DoF approaches, for example, the depth of the grasp is not directly regressed but instead uses the depth image to recover the grasp depth conditioned on the grasp position and 2-finger gripper wrist rotation~\cite{morrison2018closing}. 

\citet{sundermeyer2021contactgraspnet} reduced the 6-DoF grasping to a 4-DoF representation by ensuring one of the contact points for a two-finger parallel gripper was taken from the point cloud. The 3-DoF rotation and gripper width was then estimated. \citet{breyer2021volumetric} directly output the predicted grasp quality, orientation, and opening width for each voxel in a queried 3D volume. The 3D position is recovered from the center of a voxel. \citet{jeng2020gdn} propose a coarse-to-fine representation, where the orientation is initially coarsely discretized as a grid with a given confidence value. A refinement step is also used for each grasp pose to allow further flexibility from the discretized coarse representation. \citet{gou2021rgb} estimate an $SO(3)$ orientation and confidence of every pixel directly from an RGB image. An analytical method based on the depth image is then used to find the gripper width and position for each pixel. \citet{ni2020pointnet} directly regressed grasps from sparse point clouds. They predict the grasp quality, approaching direction, and opening direction of the gripper for every point in the original point cloud. Likewise, the approach by \citet{li2021simultaneous} estimates the rotation, width, depth, and quality of a grasp for each point in the point cloud. They combine this with another branch which predicts whether the grasp would be in collision. Alternatively, \citet{choi2018soft} predicts the most likely grasping direction and wrist orientation from a set of discretized directions and orientations. The translational part of the pose is found by determining the centered contacting voxel within the grasping direction. \citet{Wang2020} assume they are grasping the centroid of the object and attempt to directly regress the quaternion for the grasp. \citet{wu2020grasp} create a network with three branches, trained to predict the likelihood of the grasp consisting of antipodal points, grasp offset for each of the sampled $\textrm{SE}(3)$ points, and grasp confidence (trained on whether grasp was executed in simulation). \citet{qin2020s4g} trains a neural network based on \citet{qi2017pointnet} to predict both grasp point and quality for every point in a point cloud.

\paragraph{Multi-Stage Approach}

These approaches use multiple stages, where a stage refers to a component of the end-to-end model with a loss function and at least one specific task. This aims to simplify learning by breaking the problem down into smaller parts. Most multi-stage approaches consider three-stage approaches~\cite{zhao2020regnet, Wang_2021_ICCV, Li2021, zhu20216dof, fang2020graspnet}, while, \citet{wei2021gpr} propose a two-stage approach. Stages are commonly done in series (one after another), however, the last two stages in the work by \citet{fang2020graspnet} are in parallel (simultaneously). 

The first stages of the network are commonly used for at least one of the following: predicting grasp quality for subsampled points~\cite{zhao2020regnet, fang2020graspnet, Wang_2021_ICCV}, estimating a grasp for each point~\cite{wei2021gpr, ni2021learning}, estimating a subset of the DoF~\cite{fang2020graspnet} or employing contrastive loss functions~\cite{zhu20216dof}. The middle stages can be used to create grasp proposals (if they were not generated in a previous stage)~\cite{zhao2020regnet,Wang_2021_ICCV} or further estimate a subset of the DoF~\cite{zhu20216dof}. 

The last stage can act as a refinement stage to improve the regressed grasps~\cite{zhao2020regnet, wei2021gpr, ni2021learning}. \citet{fang2020graspnet} used two final parallel stages to generate grasp proposals and predict the ability of the grasp pose to tolerate larger errors, aiming to improve the robustness to imperfect sensing or control. The final stage can also be used to predict the remaining DoF~\cite{zhu20216dof}. Example grasps regressed in the work by \cite{fang2020graspnet} from real-world data are shown in \cref{fig:direct}.

An alternative multi-stage approach is demonstrated by \citet{shao2020unigrasp} and \citet{Li2021} who propose an approach that uses each stage of the network to predict a single contact point. The subsequently regressed contact point will be conditioned on the previous point. The final grasp is then recovered from the regressed contact points. Furthermore, \citet{shao2020unigrasp} shows that this approach is generalizable between different robotic hands.

\subsection{Reinforcement Learning}
\label{sec:rl}

Reinforcement Learning (RL) approaches aim to learn a policy to maximize the cumulative reward commonly over a multi-step task. We only review deep RL, where the policy is parameterized by a deep neural network. We sub-divided the reviewed works into two main approaches: On- and Off-Policy Learning. For a more comprehensive review on deep RL, see ~\cite{rl_review} and for a deeper exploration of RL in grasping and its open challenges, see \cite{grasping_rl2020review}.

Some seminal works in reinforcement learning for grasping include the work by \cite{levine2016learning, kalashnikov2018qtopt, quillen2018deep}. \citet{levine2016learning} collected 800k grasp attempts over multiple months. They present a self-supervised approach, and the authors claimed their approach is analogous to an RL formulation. \citet{kalashnikov2018qtopt} and \citet{quillen2018deep} presented works that formulate grasping as a reinforcement learning problem. \citet{kalashnikov2018qtopt} focus on real-world data collection, collecting 580k real-world grasping attempt, while \citet{quillen2018deep} train on purely simulated data. \cref{tab:rlPapers} shows and summarizes all reinforcement learning based works found during our review.

\begin{table}[tb]
\centering
\caption{Our systematic survey found 10 publications employing \textit{Off Policy} or \textit{On Policy} reinforcement learning (RL). The LfD column indicates approaches which learn from expert demonstration.}
\begin{tabular} {| c | c ? c | c | c |}
\hline
\rowcolor[HTML]{D0CECE}
Year & Paper & Learning & Algorithm & LfD \\ \hline
2018 & \citet{gualtieri2018learning} & On Policy & DQN\cite{mnih2015human} & \\ \hline
2019 & \citet{wu2019pixel} & On Policy & PPO\cite{schulman2017proximal} & \\ \hline
2019 & \citet{2019_ICRA_mbkrb} & On Policy & TRPO~\cite{trpo} & \\ \hline
2020 & \citet{mandikal2020graff} & On Policy & PPO\cite{schulman2017proximal} & \\ \hline
2020 & \citet{song2020grasping} & Off Policy & Q Learning\cite{watkins1992q} & \ForestGreencheck\\ \hline
2020 & \citet{wu2020generative} & On Policy & PPO\cite{schulman2017proximal} & \\ \hline
2021 & \citet{berscheid2021robot} & On Policy & Single Step MDP & \\ \hline
2021 & \citet{kawakami2021learning} & On Policy & PPO\cite{schulman2017proximal} & \ForestGreencheck\\ \hline
2021 & \citet{Tang2021LearningCP} & Off Policy & Q Learning\cite{watkins1992q} & \\ \hline
2021 & \citet{wang2021goalauxiliary} & Off-Policy & DDPG\cite{lillicrap2015continuous} & \ForestGreencheck\\ \hline

\end{tabular}

\label{tab:rlPapers}
\end{table}

\subsubsection{On-Policy} We found that out of the reviewed work employing RL, On-Policy methods were more common~\cite{kawakami2021learning, gualtieri2018learning, berscheid2021robot, mandikal2020graff, wang2021goalauxiliary, wu2019pixel, wu2020generative, chen2020transferable}. In On-Policy RL, training a policy is done using experiences that are collected from the most recent policy. In the work from \citet{kawakami2021learning}, the grasping task is divided into consecutive stages: orienting the end-effector, approaching the target, and closing the gripper. A different RL model is trained for each stage, and curriculum learning is employed that adjusts the reward function based on the success rate of each task. \citet{gualtieri2018learning} used a Deep Q-Network~\cite{mnih2015human} with Monte Carlo updates to learn how to grasp an object and place it into a desired configuration. Working with a 24-DoF hand, \citet{mandikal2020graff} trained an actor-critic model. They proposed a two-step architecture: Initially, a CNN, which is trained on ContactDB~\cite{brahmbhatt2019contactdb}, estimates the pixel regions that belong to a ``use" affordance. An RL policy then takes this affordance mask, RGB-D image, and gripper configuration as input and outputs the wrist pose and the 24-DoF robot hand configuration. A sparse reward is awarded when the object is lifted from the ground, and a dense reward is awarded according to a distance metric from the affordance region. \citet{chen2020transferable} used an advantage actor-critic policy gradient to train a policy that will optimize the viewpoint for grasping. They then apply the grasping algorithm developed by \citet{pas2015using} to calculate grasps for the optimized viewpoint. \citet{wu2019pixel} used a single depth image and introduces a novel attention mechanism that learns to focus on sub-regions of the depth image in order to grasp better in cluttered environments. They formulate the problem using a policy gradient method based on PPO~\cite{schulman2017proximal}. \citet{wu2020generative} extended this framework to robotic hands with arbitrary degrees of freedom. \citet{2019_ICRA_mbkrb} propose the use of TRPO to learn control policies that take contact feedback as input. They show that this policy significant improved the robustness of the grasp under both object pose uncertainty and shape complexity.

\subsubsection{Off-Policy}
Off-Policy RL methods use the data collected throughout training to train a new policy. Employing human demonstrations, \citet{song2020grasping} used Q-learning to estimate the optimal Q-function. They simulate future states by giving an action for the robot to complete, allowing the algorithm to forward simulate possible future states conditioned on the current state-action pair. \citet{Tang2021LearningCP} demonstrated collaborative pushing actions to facilitate grasping. Their approach uses Q-learning to learn a deterministic policy for pushing and grasping. No reward is assigned to pushing actions - the agent is only rewarded when the robot successfully grasps the object. \citet{berscheid2021robot} formulated the problem as a Markov Decision Process with a single action step. They train a fully convolutional neural network to learn a 4-DoF planar grasping system. A model-based controller decides the other two degrees of freedom by avoiding collisions and maximizing grasp quality. \citet{wang2021goalauxiliary} trained a grasping policy from demonstrations based on the Deep Deterministic Policy Gradient algorithm~\cite{lillicrap2015continuous}. The demonstrations are obtained using a motion and grasp planner, which is assumed to be an `expert' in their formulation. 

\subsection{Exemplar Methods}
\label{sec:other}

Exemplar methods attempt to transfer grasps from previous examples. \citet{patten2020dgcm} designed an approach to grasp novel objects by learning from experience. This is achieved using metric learning to encode objects with similar geometries nearby in feature space. Finding a successful grasp is framed as a nearest neighbor search through feature space, searching for a previously successful grasp. The approach by \citet{Dexnet1} compared a given grasp candidate against a database of successful grasps. They compare grasp candidates using three feature maps: grasp parameters, depth match gradient at local patches around the object, and similarity of the object model assessed by a deep-learning based network. The feature maps form a prior belief distribution on the similarity to all grasps in the database. The grasp with the maximum lower confidence bound of the belief distribution is executed.

\section{Supporting Methods Based on Deep Learning}

Deep-learning can be applied to methods throughout the grasping pipeline that aim to improve the success of a grasping task. This task does not have to focus solely on picking up an object. More complex manipulation tasks may require grasping that affords a particular subsequent action. For example, when handing over a full mug, the robot may need to grasp the mug handle. We found two clusters of supporting methods in the reviewed literature: shape approximation techniques and affordances based methodologies.

\subsection{Shape Approximation}
\label{sec:shape}

The most common form to approximate the shape of an object from partial information found in the literature is \textit{Shape Completion}, which aims to estimate the full object model from a partial input shape (e.g. point cloud of an object from one camera view). We define \textit{Shape approximation} more generally to include any method which approximates shape from an input. This includes the approximation of the actual shape of an object by simple(r) shapes and the fusion of multimodal data to approximate a shape.

\subsubsection{Shape Completion}
\citet{varley2017shape} trained a 3D CNN to employ shape completion on a single view voxel grid, outputting a voxel grid with shape completed object. \citet{gao2019kpamsc} instead use the method from \citet{zhang2018learning} for shape completion, predicting a 3D voxel grid directly from RGB-D images. \citet{Kiatos2020} use a variational autoencoder~\cite{Kingma_2019} to predict the occluded surface points and associated normals of a partial 3D point cloud. \citet{chavandafle2021object} predicted the depth image that estimates the `back' side of an object from a masked depth image. The front and back sides can then be stitched together quickly to form an object mesh.  

\begin{figure}
    \centering
    \includegraphics[width=0.9\linewidth, trim={0 2cm 13cm 0}, clip]{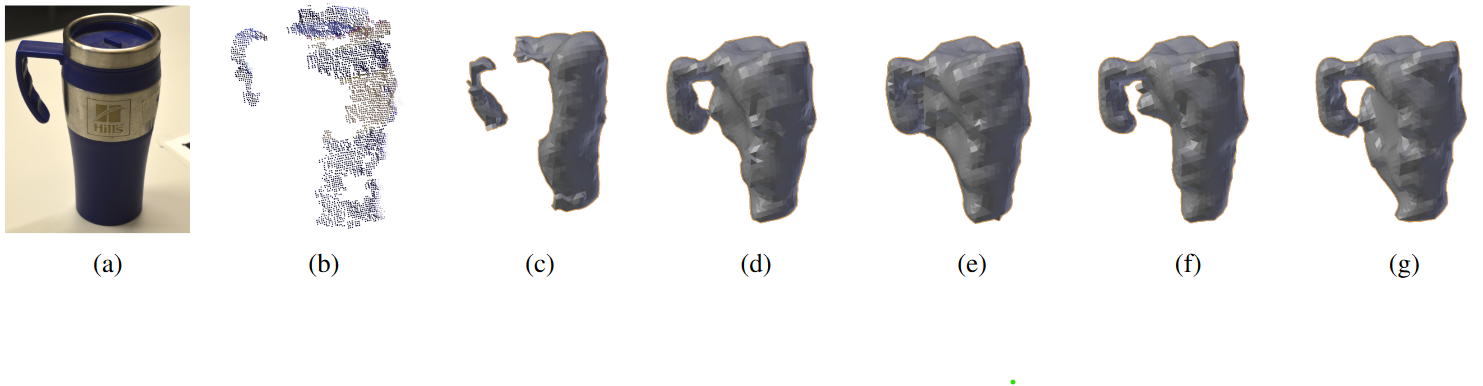}
    \caption{\citet{Lundell2019} uses drop out layers to generate 20 shape completed samples, where the average of these samples is shown in (d) compared to~\cite{varley2017shape}(c). ($\copyright$2019 IEEE)}
    \label{fig:sm_shape_completion}
\end{figure}

The uncertainty around the output of the shape completion can be useful. \citet{gualtieri2021robotic} incorporate uncertainty in their shape completion network, where it represents the estimated probability that each predicted point is accurate. Another approach including uncertainty is \citet{Lundell2019} which incorporates a Monte-Carlo drop-out procedure~\cite{gal2016dropout} to generate a series of shape completed objects. GraspIt!~\cite{GraspIt} is then used to plan grasps over the mean object shape. The most suitable grasp over the series of shapes is chosen as the grasp point. An example of this procedure is shown in \cref{fig:sm_shape_completion}. Interestingly, uncertainty is ignored in follow-on work from the same authors~\cite{lundell2020beyond, lundell2021ddgc, lundell2021multifingan}. 

\subsubsection{Auxiliary Tasks}
In deep learning, auxiliary tasks can be added into a deep-learning model, which has been shown to boost the performance of the model in some domains~\cite{vafaeikia2020brief}. \citet{jiang2021synergies} asserted that 3D reconstruction and grasping are closely related, where both rely on knowledge of an object's local geometries. Authors have proposed learn object reconstruction as an auxiliary task to grasping~\cite{jiang2021synergies, yang2020robotic,yan2018learning}. \citet{jiang2021synergies} used a self-supervised approach to reconstruct an object and calculate a grasp. \citet{yang2020robotic} simultaneously regressed a grasp pose and reconstructed the point cloud of an object. The grasp pose is then refined by projecting it onto the surface of the point cloud reconstruction. \citet{yan2018learning} employed two networks, one for shape completion and another for grasp outcome prediction. They demonstrated a performance improvement when the grasping network uses the feature space representation produced by the shape generation network.

\subsubsection{Other}

\citet{avigal20206dof} do not complete explicit shape prediction, however, they used a network that takes RGB images from multiple viewpoints and generates the corresponding depth maps for the shape. The depth maps are then fed into a 4-DoF grasping algorithm~\cite{mahler2017dexnet}. \citet{ottenhaus2019visuo} used Gaussian Process Implicit Surfaces~\cite{Bjrkman2013EnhancingVP} to fuse visual and tactile sensor inputs. After capturing a point cloud of the object, the robot gathers information about unseen sides of objects using tactile information. Researchers have shown that many everyday objects can be modeled as simple shapes~\cite{Shiraki2014,Miller2003}. Using this observation, \citet{Torii2018} approximated objects as a series of 3D primitive shapes (hexahedron, cylinder, sphere). They use a neural network to predict the likelihood of each primitive shape before using a pre-computed database of rules to perform the grasping. 

\subsection{Affordances}

\label{sec:semantics}

This section reviews how affordances have been used in the domain of 6-DoF grasping. See \cite{ardon2020survey} for a review of affordances in the more general robotics domain. In addition to considering the success of grasping, these approaches have additional considerations for what kind of task it is used for. For example, when a \emph{handover} task is to be completed for a scissors object, the robot should grasp the scissors by the blade and then pass it over to the human who can grasp it by the handle. The robot's understanding of how an object is used in a particular task or how humans use those objects can lead to higher-level reasoning about the grasping task. This lends itself towards 6-DoF grasping, as different parts of the object need to be grasped in different ways, depending on location of the affordance.

\begin{figure}
    \centering
    \includegraphics[width=0.6\linewidth, trim={0cm 0cm 38cm 11.8cm}, clip]{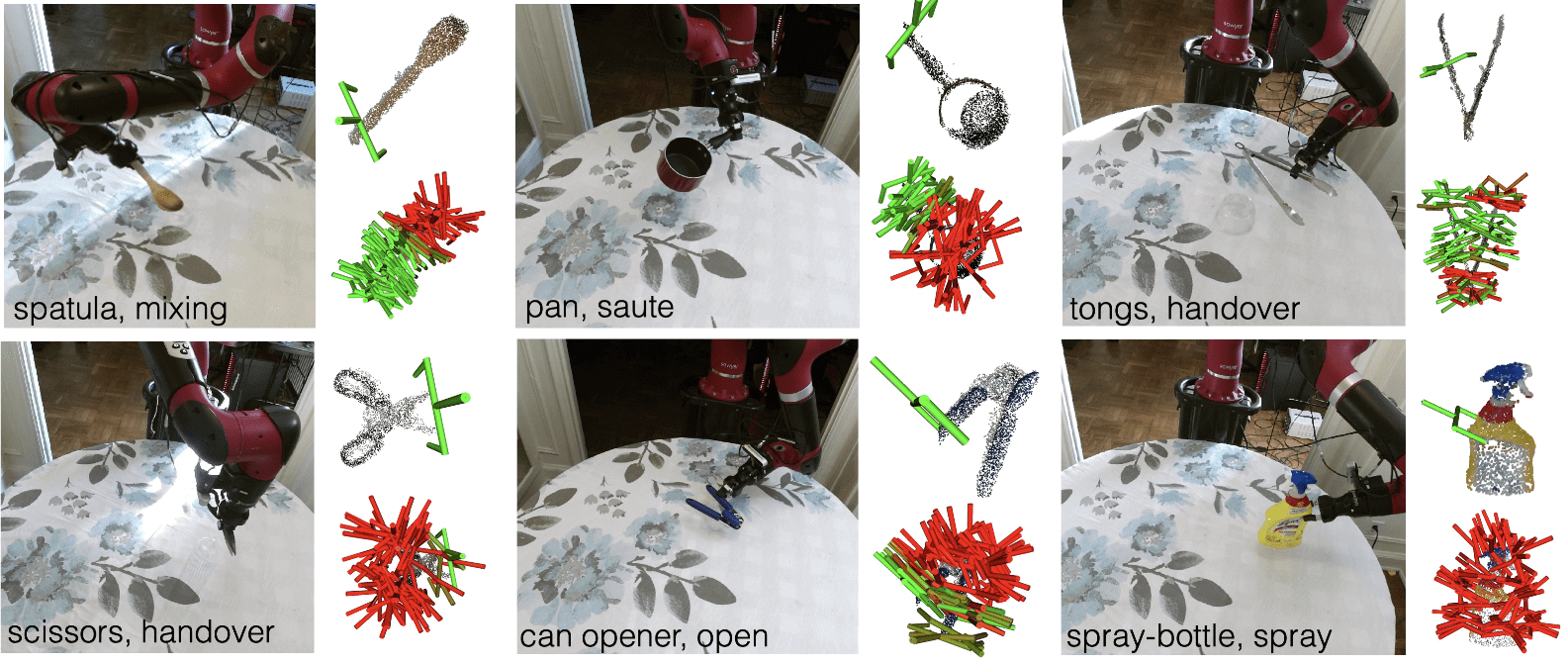}
    \caption{~\citet{murali2020same} studies task-oriented grasps on unknown objects. The top right visualization shows the grasp with the highest quality considering a given affordance. The bottom right shows all the stable grasp candidates, colored by the relevance to the affordance (green is high)}
    \label{fig:sm_semantics}
\end{figure}

Some researchers use deep neural networks to segment objects for different affordances~\cite{li2020learning, do2018affordance, mandikal2020graff, nguyen2016detecting, Nguyen2017}. They then use analytical methods to find grasps within the segmented object portion with an appropriate affordance. Alternatively, the approach by \citet{murali2020same} estimated the quality of a sampled grasp given an affordance label. This is shown in \cref{fig:sm_semantics}, where the grasps shown in green are relevant to the given affordance. \citet{Ardon2019} created a knowledge base graph representation using Markov Logic Networks to obtain a probability distribution of grasp affordances. Additionally, both \citet{manuelli2019kpam} and \citet{gao2019kpamsc} presented an approach based on detecting a fixed number of keypoints for a category of objects.

\section{Dataset Design}
\label{sec:training}

\subsection{Objects Sets}
\label{sec:objects}

The objects used for training and testing grasping algorithms are crucial for the reported grasp success and allowing the community to reproduce the results. Researchers commonly use subsets of existing object sets when investigating grasping. However, there is no standard procedure for selecting this subset. This can lead to inconsistencies in objects between different works which are using the same object set. This increases the difficulty of comparing grasping performance between works. The most commonly used object set in the reviewed works is Yale-CMU-Berkeley(YCB)~\cite{calli2015ycb}, being used almost twice as often than the next most adopted object set (\cref{tab:objectsets}). 

\begin{figure}[bt]
    \centering
    \includegraphics[width=0.7\linewidth]{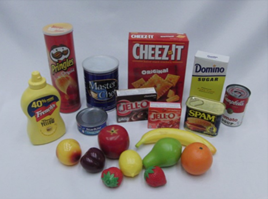}
\caption{A commonly used object set in robotic grasping is the YCB object set~\cite{calli2015ycb}. It consists of a set of daily household objects, with a subset of the objects from the `Food' category shown here. ($\copyright$2015 IEEE) 
} 
\label{fig:ycb}

\end{figure}

YCB~\cite{calli2015ycb} (shown in \cref{fig:ycb}), BigBIRD~\cite{singh_bigbird_2014}, KIT~\cite{kit2012} and Cornell~\cite{lenz2014deep} consist of mostly household items such as food, toys and tools. These object sets are appropriate for service robotics, but may not test the robustness of grasping algorithms on complex objects. ShapeNet~\cite{shapenet2015}, 3DNet~\cite{wohlkinger_3dnet_2012}, Grasp~\cite{bohg2015dataset}, PSB~\cite{shilane_princeton_2004}, ModelNet~\cite{zhirong_wu_3d_2015}, ObjectNet3D~\cite{xiang_objectnet3d_2016} and ContactDB~\cite{brahmbhatt2019contactdb} include object model repositories, containing a large number of virtual object models, mainly used for training and testing in simulation. 

EGAD!~\cite{morrison2020egad} and Procedural~\cite{bousmalis2018grasping} consist of procedurally generated object models. EGAD!~\cite{morrison2020egad} proposes a set of 3D printable objects that vary in terms of grasping difficulty~\cite{mahler2017dexnet} and object complexity~\cite{Wang75}. Procedural~\cite{bousmalis2018grasping} generate a simulated object set by attaching rectangular prisms in random orientations and locations.

Most real-world object sets do not provide a standardized method to acquire the physical objects consistently (an exception to this is YCB\footnote{https://www.ycbbenchmarks.com/}). One solution to this is 3D printed datasets such as EGAD!\footnote{https://dougsm.github.io/egad/}, however, these objects lack semantic meaning.

\begin{table}
\centering
\caption{Number of times objects sets are used in reviewed papers.}
\begin{tabular} {| c | c | c |}
\hline
\rowcolor[HTML]{D0CECE}
Object Set/Database & \# of times used & Sim//Real\\ \hline
YCB~\cite{calli2015ycb} & 29 & Real \\ \hline
3DNET~\cite{wohlkinger_3dnet_2012} & 15 & Sim\\ \hline
BigBIRD~\cite{singh_bigbird_2014} & 12 & Real\\ \hline
KIT~\cite{kit2012} & 12 & Real\\ \hline
ShapeNet~\cite{shapenet2015} & 11 & Sim\\ \hline
Grasp~\cite{bohg2015dataset}  & 10 & Sim\\ \hline
EGAD!~\cite{morrison2020egad} & 2 & Sim/Real \\ \hline
Cornell~\cite{lenz2014deep} & 2 & Real\\ \hline
Dex-Net~\cite{mahler2017dexnet} & 2 & Real\\ \hline
PSB~\cite{shilane_princeton_2004} & 1 & Sim\\ \hline
ModelNet~\cite{zhirong_wu_3d_2015} & 1 & Sim\\ \hline
ObjectNet3D~\cite{xiang_objectnet3d_2016} & 1 & Sim\\ \hline
ContactDB~\cite{brahmbhatt2019contactdb} & 1 & Sim\\ \hline
Procedural~\cite{bousmalis2018grasping} & 1 & Sim\\ \hline
Custom & 11 & Real/Sim\\ \hline
\end{tabular}
\vspace{0.2cm}

\label{tab:objectsets}
\end{table}

\subsection{Procedurally Generated Datasets}

Even though a majority of the reviewed works use benchmark object sets, they commonly opt to create their own custom datasets with those objects. A number of works used datasets collected using real robots~\cite{choi2018soft,gou2021rgb,berscheid2021robot,Wang2021,kasaei2021mvgrasp} and some combined simulation and real-world data~\cite{fang2020graspnet,gou2021rgb,Wang2021}. However, the large majority opted to use purely simulated datasets when training their networks.

Some authors have released their datasets, with reviewed works employing public datasets such as GraspNet-1Billion~\cite{fang2020graspnet}, a hybrid 6-DoF grasping dataset that captures real RGB-D camera data and combines this with simulated grasp poses, Shape Completion Grasping~\cite{varley2017shape}, a database of voxel grid pairs for shape completion and ACRONYM~\cite{eppner2020acronym}, a simulation-based dataset for 6-DoF grasping.

\subsection{Expert Datasets}

\begin{figure}[tb]
    \centering
    \includegraphics[width=0.7\linewidth]{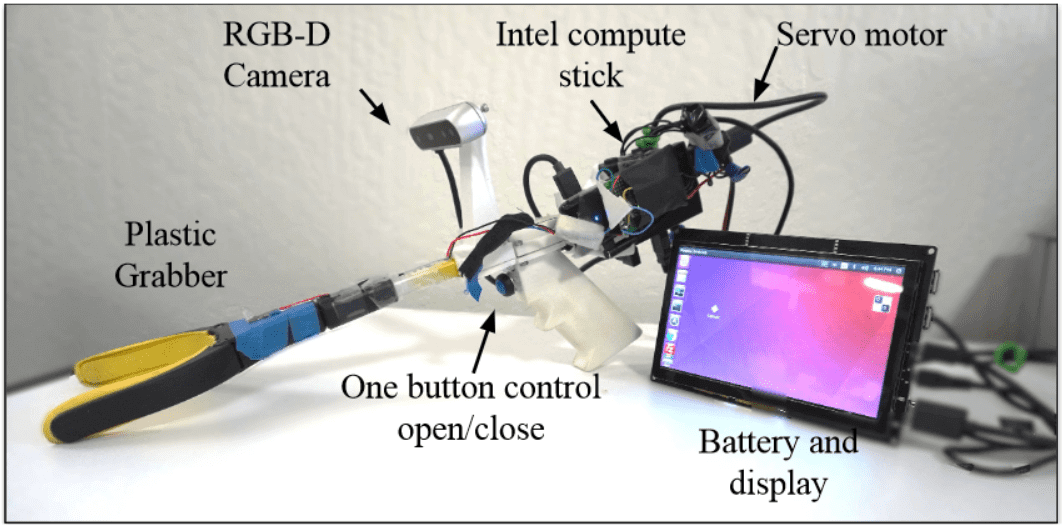}
    \caption{\citet{song2020grasping} design a low-cost handheld gripper to generate human annotations for grasping. ($\copyright$2020 IEEE)}
    \label{fig:lfd}
\end{figure}

Researchers have proposed various datasets which consist of expert demonstrations, either from a human or an algorithm. \citet{yan2018learning} generates a dataset of around 1.6k human grasping demonstrations within Virtual Reality (VR) from 5 people. Also in VR, \citet{kawakami2021learning} creates a system to collect grasping demonstrations. The operator demonstrates the position of the arm using a controller with tracked position and pose. Using a handheld gripper (see \cref{fig:lfd}), \citet{song2020grasping} generated a dataset from human demonstration. The dataset contains 12 hours of gripper-centric RGB-D videos, with each picking attempt separated into short clips to correspond to grasps. Similarly, \citet{Cortes2021} generated 300 demonstrations by kinesthetically teaching the robot trajectories. \citet{osario2020human} created a dataset of human grasps where the human is controlling a robotic gripper with a joystick. Alternatively, the approach by 
\citet{wang2021goalauxiliary} used demonstration from Optimization-based Motion and Grasp Planner (OMG planner)~\cite{OMG2020}. \citet{GRAB:2020} presented a dataset of whole-body grasps generated by ten subjects interacting with 51 everyday objects of varying shape and size. The object meshes are annotated with the contacts created by the human hand.

\subsection{Data Representation}

There are four major sensor inputs used for the deep learning methods: Point Cloud; Voxel Grid; RGB-D Image; and Depth Image. These four representations are all interchangeable for spatial data, assuming that intrinsic parameters for the camera are known. \cref{tab:input} shows the popularity of each of the input formats and \cref{tab:backbone} shows the popularity of different network architectures.

\begin{table}[tb]
\centering
\caption{Number of times each input format has been used}
\begin{tabular} {| c | c |}
\hline
\rowcolor[HTML]{D0CECE}
Input Format & Number of times used\\ \hline
Point Cloud & 34 \\ \hline
Depth Image & 18 \\ \hline
RGB-D Image & 15 \\ \hline
Voxel Grid & 12 \\ \hline
Segmentation Mask & 9 \\ \hline
Other & 10 \\ \hline
\end{tabular}

\label{tab:input}
\end{table}

\textbf{Point Cloud}: Point clouds are the most popular data format, especially with the advancements in networks that can learn directly from point clouds, such as PointNet~\cite{Charles2017}. In addition, the point cloud representation allows researchers to easily fuse data from multiple viewpoints if the relative camera poses are known. 

For point cloud based data, researchers commonly used PointNet~\cite{Charles2017} and PointNet++~\cite{qi2017pointnet}. This network backbone type has been used in different methods including direct regression~\cite{fang2020graspnet, wu2020grasp, ni2020pointnet, qin2020s4g, sundermeyer2021contactgraspnet, jeng2020gdn, zhao2020regnet}, RL~\cite{wang2021goalauxiliary}, sampling~\cite{Liang_2019,peng2021selfsupervised,mousavian20196,murali20206,murali2020same}, and shape completion~\cite{gualtieri2021robotic}. 


Point cloud for direct regression tends to subsample points and learn grasps for them~\cite{fang2020graspnet, qin2020s4g, zhao2020regnet, ni2020pointnet, ni2021learning, Kokic2020, Wang2021, wei2021gpr, li2021simultaneous, jeng2020gdn, Li2021, Wang2020}. When a point cloud is used for sampling, one of the following procedures is usually followed. The points inside the gripper are fed to the network~\cite{Ren2021, Liang_2019, corsaro2021learning}, the points will be transformed such that the origin aligns with the grasp frame~\cite{corsaro2021learning, wencatgrasp}, or a representation of the gripper will be rendered, and all of the points will be fed into the network~\cite{mousavian20196}. Minimal surveyed work has investigated the use of point clouds in RL apart from ~\cite{wang2021goalauxiliary}.

\textbf{Images}: The next most popular formats are depth and RGB-D images. Learning from images has been highly studied in both computer vision and robotics, allowing researchers to train pre-existing architectures for robotic grasp synthesis. 

For images, the most common approach is to employ ResNet~\cite{ResNet} as the backbone~\cite{song2020grasping,gou2021rgb,yang2020robotic,zhu20216dof,liu2020deep,lundell2021ddgc,patten2020dgcm,lundell2021multifingan,bajracharya2020mobile}. In addition to ResNet, various other architectures have been used, including VGG~\cite{simonyan2015deep} LeNet~\cite{LeNet}, DenseNet~\cite{DenseNet} and U-Net~\cite{ronneberger2015unet}. 

Direct regression approaches commonly use the whole image as input~\cite{gou2021rgb, zhu20216dof, Schmidt2018}. For sampling, some approaches generated a depth image slice representing points within the gripper~\cite{ten2017grasp, gualtieri2017high} while others concatenated grasp features after processing images through a CNN~\cite{lu2019planning, lundell2021multifingan, lundell2021ddgc, aktas2019deep}. RL approaches input the current camera view to generate an action~\cite{song2020grasping, berscheid2021robot, wu2019pixel, wu2020generative, gualtieri2018learning, Tang2021LearningCP, Cortes2021}.

\textbf{Voxel Grids}: A voxel represents a value in a regular 3D grid. Voxel Grids are analogous to images in 3D space, where a voxel is similar to a pixel over a 3D grid instead of a 2D image. 

For voxel based inputs, VoxNet~\cite{VoxNet} is most commonly used. VoxNet integrates a volumetric occupancy grid representation of the data with a 3D-CNN. Voxel grids have been used for shape completion~\cite{varley2017shape, Kiatos2020}, sampling~\cite{lou2020learning, lu2020multi, lu2020multifingered} and direct regression~\cite{liu2019generating,choi2018soft, breyer2021volumetric}. For sampling, Voxel grids have been used similarly to point clouds, transforming the origin to align with grasp frame coordinates~\cite{lou2020learning} or adding grasp configuration features after the convolutional layers~\cite{lu2020multi, lu2020multifingered, ottenhaus2019visuo}. Voxel grids have also been used to directly regress a single pose using a subset of the voxel grid corresponding to the object~\cite{liu2019generating, choi2018soft} as well as directly regressing a grasp for each voxel~\cite{breyer2021volumetric}.

\begin{table}
\centering
\caption{Number of times each network backbone has been used. Some papers use multiple backbones, depending on their network architecture.}
\begin{tabular} {| l | c |}
\hline
\rowcolor[HTML]{D0CECE}
Network Backbone & Number of times used\\ \hline
PointNet/PointNet++~\cite{qi2017pointnet} & 21 \\ \hline
ResNet~\cite{ResNet} & 9 \\ \hline
LeNet~\cite{LeNet} & 3 \\ \hline
Other & 20 \\ \hline
Custom & 31 \\ \hline
\end{tabular}
\label{tab:backbone}
\end{table}

\section{Benchmarking}
\label{sec:benchmarking}

\subsection{Experimental Evaluation}

Most approaches implemented their grasping method in the real-world, either as an evaluation of the performance with respect to the metrics described in Section \cref{sec:metrics} or presenting a demonstration that acts as a proof-of-concept of their method. Demonstrations focus on showing that their system works in the real-world without systematically evaluating their method. However, some approaches only consider evaluating their system using a simulator.

Evaluating in the real-world naturally carries more weight since the goal for robotic grasping is to be ultimately applied to real world. It should be noted that even though most works train their models using purely simulated data, they evaluate their approach in real world experiments. Commonly, the grasping system was directly transferred from simulation to real-world (e.g \cite{mousavian20196, fang2020graspnet, ten2017grasp, gualtieri2017high, varley2017shape, wu2020grasp}). Techniques such as domain adaptation~\cite{domaingrasping} or domain randomization~\cite{random1, random2} have also been used to transfer grasping approaches between simulation and real-world. \citet{zhu20216dof} used contrastive learning~\cite{contrastive} to extract invariant features when images are augmented, aiming to improve the model under image sensor noise.

\subsection{Hardware}

\begin{table}[tb]
\centering
\caption{The popularity of various types of hardware systems used throughout the surveyed papers. The most popular category of robotic arms is further analyzed in \cref{tab:rob_arms}.}
\begin{tabular}{|l|c|}
\hline
\rowcolor[HTML]{D0CECE}
Category & Popularity \\ \hline
Robot Arm  & 66 \\ \hline
Humanoid   & 6  \\ \hline
Mobile Arm & 3  \\ \hline
Not Used   & 7  \\ \hline
\end{tabular}

\end{table}

We found that most works that study 6-DoF grasping use a robotic arm. However, a minority of works use other platforms, such as mobile platforms or humanoid robots. The robotic arms were most commonly table-mounted, and the robot would perform table-top grasping. Humanoid robots were commonly used when studying papers relating to affordances~\cite{Schmidt2018,nguyen2016detecting,Nguyen2017}. Surveyed works also completed research using mobile robots for grasping, however, none of the works made use of the extra DoF unique to the mobile aspect of the robot.


\begin{table}[tb]
\centering
\caption{Popularity of various robotic arm manufacturers throughout the surveyed papers.}

\begin{tabular}{|l|l|c|c|}
\hline
\rowcolor[HTML]{D0CECE}
Manfacturer        & Robot    & \multicolumn{1}{l}{Popularity} & DoF \\ \hline
\rowcolor{gray_table}
Franka Emika       & Panda    & 15                             & 7   \\ \hline
\rowcolor{white}
Universal Robotics & UR5      & 9                              & 6   \\
\rowcolor{white}
Universal Robotics & UR10     & 4                              & 6   \\ 
\rowcolor{white}
Universal Robotics & UR5e     & 1                              & 6   \\ \hline
\rowcolor{gray_table}
Kuka               & LBR4     & 6                              & 7   \\
\rowcolor{gray_table}
Kuka               & IIWA LBR & 3                              & 7   \\ \hline
\rowcolor{white}
Kinova             & Gen3     & 2                              & 7   \\
\rowcolor{white}
Kinova             & Mico2    & 2                              & 6   \\
\rowcolor{white}
Kinova             & Jaco2    & 2                              & 6   \\ \hline
\rowcolor{gray_table}
Rethink Robotics   & Baxter   & 6                              & 7   \\ \hline
\rowcolor{white}
Staubli            & TX60     & 4                              & 6   \\ \hline
\rowcolor{gray_table}
ABB                & Yumi     & 2                              & 7   \\ \hline
\rowcolor{white}
                  & Custom    & 2                             &     \\ \hline
\rowcolor{gray_table}
                  & Other   & 22                              &     \\ \hline
\rowcolor{white}
                  & Not Used & 7                              &     \\ \hline

\end{tabular}
\label{tab:rob_arms}
\end{table}

Most researchers used either industrial robots or robots designed for human-robot interaction. The most common robotic platform is the Franka Emika Panda. This robot has a redundant DoF which allows more freedom in joint angles when achieving a specific gripper pose.


\begin{table}[tb]
\centering
\caption{The popularity of various grippers in the surveyed works, grouped by manufacturer.}

\begin{tabular}{|l|l|c|c|}
\hline
\rowcolor[HTML]{D0CECE}
Manufacturer & Model & Popularity & DoF \\ \hline
\rowcolor{gray_table}
Robotiq & 2F & 10 & 1 \\
\rowcolor{gray_table}
Robotiq & 3F & 5 & 1 \\ \hline
\rowcolor{white}
Franka Emika & Panda Gripper & 12 & 1 \\ \hline
\rowcolor{gray_table}
Barret & BarretHand & 7 & 5 \\ \hline
\rowcolor{white}
Kinova & 2F & 3 & 1 \\
\rowcolor{white}
Kinova & 3F & 1 & 2 \\ \hline
\rowcolor{gray_table}
Wonik Robotics & Allegro Hand & 4 & 16 \\ \hline
\rowcolor{white}
Rethink Robotics & Baxter Gripper & 4 & 1 \\ \hline
\rowcolor{gray_table}
Shadow Robot Company & Shadow 
Hand & 2 & 20 \\ \hline
\rowcolor{white}
& Custom & 2 & \\ \hline
\rowcolor{gray_table}
& Other & 32 & \\ \hline
\end{tabular}
\label{tab:grippers}
\end{table}

The most commonly used gripper was a two-finger parallel jaw ($51$ times) with researchers using the gripper that came with the robotic arm. Some researchers use grippers with more than two fingers, including the Barret Hand, Allegro Hand, Shadow Hand and, Kinova 3-Finger gripper. Some papers only make use of two-fingers from the Barret hand\cite{wu2020generative} or switch between two- and three-fingers\cite{Wang2020}. Multi-fingered high-DoF hands present a different set of challenges compared to a two-finger gripper. Research on multi-fingered grippers focus on how to generate a grasp pose that considers the large amount of DoF (e.g \cite{liu2019generating, lu2020multifingered, liu2020deep, corsaro2021learning}), or grasping with affordances (e.g \cite{nguyen2016detecting,Nguyen2017,mandikal2020graff}). Soft grippers are another potentially interesting line of work with only a few reviewed works making use of them~\cite{choi2018soft,Goncalves2019,Cortes2021}.

\subsection{Performance Metrics}
\label{sec:metrics}

A diverse set of performance metrics is used among all these works. The common definitions of performance metrics related to grasping from the reviewed work are listed below and \cref{tab:performamce_metrics} shows the frequency each metric is used. However, the exact definition of each performance metric can vary slightly between different works.

\begin{enumerate}
    \item \textbf{Grasp Success Rate}: The percentage of successful grasps (No. of Successful Grasps / Total No. of Attempted Grasps). The post-grasp steps prior to considering a grasp `successful' was not consistent across reviewed works. 
    \item \textbf{Completion / Clearance Rate}: The percentage of objects that are removed from the clutter (No. of Objects Grasped / Total No. of Objects in Clutter).
    \item \textbf{Computation Time}: Time required to compute grasp hypothesis generation.
    \item \textbf{Precision}: The percentage of true positive grasp predictions (No. of True Positive Grasp Predictions / (No. of Selected Positive Grasps)).
    \item \textbf{Coverage}: The percentage of sampled ground truth grasps that are within a threshold distance of any of the generated grasps.
    \item \textbf{Grasp Prediction Accuracy}: The percentage of grasps outcomes correctly predicted (No. of Successful Grasp Predictions / No. of Predictions).
\end{enumerate}

\begin{table}[]
\caption{The frequency of common metrics employed in the reviewed works as a percentage. Success rate is the most commonly used metric throughout all reviewed papers.}
\centering
\begin{tabular}{|l|c|}
\hline
\rowcolor[HTML]{D0CECE}
Metric                    & Frequency Used (\%) \\ \hline
Success Rate              & 86          \\ \hline
Completion                & 25           \\ \hline
Computation Time           & 21          \\ \hline
Precision                 & 12           \\ \hline
Coverage                  & 6           \\ \hline
Grasp Prediction Accuracy & 8          \\ \hline
\end{tabular}
\label{tab:performamce_metrics}
\end{table}

\subsection{Object Configurations}

Object configurations are how the objects are arranged in the scene during training or testing. We cluster object configurations into three types. 

\begin{figure}[bt]
\hfill
\begin{subfigure}{0.45\linewidth}
\includegraphics[height=0.7\linewidth]{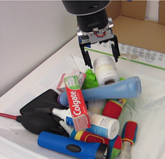}
\end{subfigure}
\begin{subfigure}{0.45\linewidth}
\includegraphics[height=0.7\linewidth]{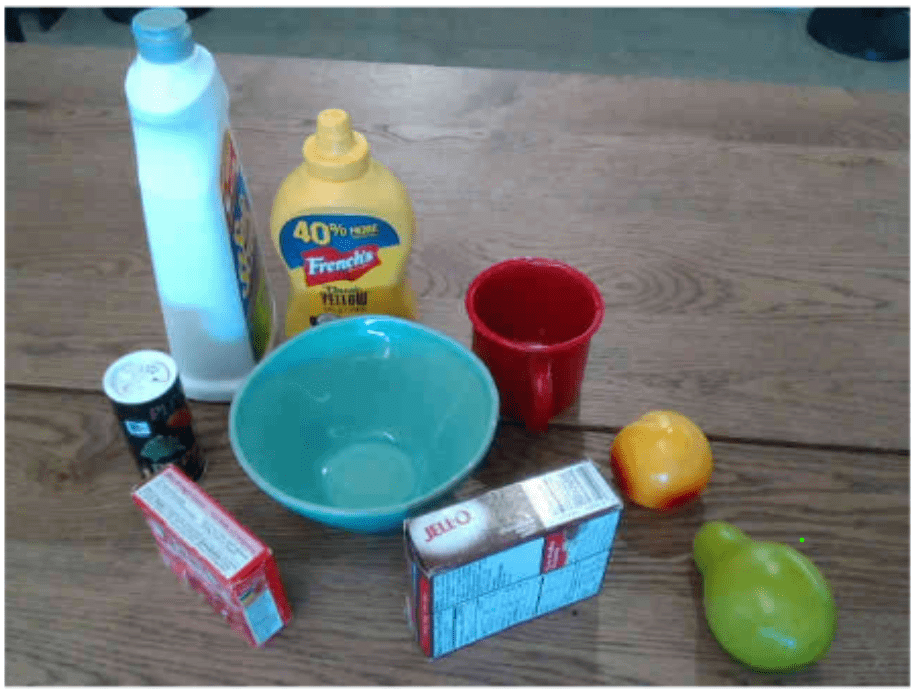}
\end{subfigure}
\hfill
\caption{Examples of cluttered scenes. We differentiate Piled Clutter (left) from Structured clutter (right). Left image is from~\cite{ten2017grasp} ($\copyright$2017 SAGE) and right image is from~\cite{murali20206}. ($\copyright$2020 IEEE)} 
\label{fig:clutter}
\end{figure}

\begin{enumerate}
    \item \textbf{Singulated}: A single object in the scene.
    \item \textbf{Piled Clutter}: Objects are packed together tightly. Objects are commonly arranged as a pile, for example \cref{fig:clutter} (Left).
    \item \textbf{Structured Clutter}: Multiple objects spread out in a scene such that they are not touching, for example \cref{fig:clutter} (Right).
\end{enumerate}

Piled clutter is frequently encountered in bin picking applications where a variety of items are compactly arranged in a bin. However, this survey only encompassed a limited number of papers which address bin picking scenarios~\cite{ten2017grasp, gualtieri2017high, song2020grasping, lou2021collisionaware, berscheid2021robot, wei2021gpr}. In bin-picking scenarios, additional constraints are needed for how to approach objects to avoid the bin's walls, making it a more challenging task than table-top scenarios.

Most works surveyed in this study which focus on singulated objects tend to be grasping with a high DOF hand~\cite{varley2017shape,vandermerwe2020learning,varley2015generating,Lundell2019,lundell2021multifingan,liu2019generating,lu2020multi,lu2020multifingered,liu2020deep,ottenhaus2019visuo,Schmidt2018,Kokic2020,aktas2019deep,Li2021}. These papers tend to focus on solving a specific grasping task, rather then aiming for generalized grasping. Other works on singulated objects focus on data representation and the learning process~\cite{wu2020grasp, zhou20176dof, wang2021goalauxiliary, patten2020dgcm}. Singulated object grasping is also used in other contexts such as affordances~\cite{murali2020same, mandikal2020graff, Ardon2019} or manipulation~\cite{gao2019kpamsc}. Researchers generally do not distinguish between structured or piled clutter scenes, however, we consider these distinct scenarios, which may have different solutions.

\section{Discussion and Future Directions}
\label{sec:future}

This section discusses the state of the field of deep learning-based grasp synthesis and highlights recommendations for future research directions. We note the key takeaways of each subsection indicated as \textbf{Key Takeaways}.

\subsection{Deep Learning Methods}

There is currently no consensus on when to use direct regression-based methods versus sampling-based methods, as both approaches focus on similar tasks with similar success rates.

Direct regression-based approaches have demonstrated the ability to run in real-time~\cite{sundermeyer2021contactgraspnet, breyer2021volumetric}, but processing speeds can vary widely depending on the chosen architecture. In contrast, sampling-based methods offer the advantage of adjustable processing speed, as the number of samples or the level of optimization can be easily modified. However, based on our review of the literature, no sampling approach has been found to have real-time capabilities.

Depending on the use case of a sampling system, different sampling methods have different advantages. Euclidean sampling tends to quickly generate a wide array of grasps of varying quality. However, a large number of grasps are needed to try to find a high-quality grasp. In contrast, sampling through priors or latent space~\cite{mousavian20196} tends to sample higher quality grasps, however, can come at a cost of time to create the samples.

RL approaches for grasping typically focus not only on grasp generation but also on the trajectory used by the system. However, RL approaches typically generate only one grasp per scene, making it difficult to incorporate additional constraints during execution.

Exemplar methods, which rely on a knowledge database of objects with similar shapes, were the least commonly used approach in our literature review. This is likely because they are limited in their ability to effectively grasp objects that differ significantly from those in the database.

\subsection{Benchmarking}

Grasping in complex environments has had recent attention in adjacent fields such as agriculture\cite{agri_review}, whereas in pure grasping literature, authors typically focus on grasping objects from tabletop scenes where most of the synthesized grasp poses are kinematically feasible. Studying 6-DoF grasping in enclosed spaces such as shelves~\cite{Costanzo2021}, around obstacles~\cite{lou2020learning,lou2021collisionaware}, around humans~\cite{Wang2021}, near reachability limits~\cite{lou2020learning} or in-the-wild (e.g orchards~\cite{WANG2022106716}) would pose more constraints on the synthesis of the grasp configurations and require the grasp poses to not only be of high quality but also diverse, to increase the probability of finding a feasible trajectory to reach those grasp poses. As 6-DoF grasping approaches are not often tested in such challenging environments or compared directly to 4-DoF ones, the question of whether to use a 6 DoF or a 4-DoF grasping approach for a given application is not adequately answered by the current state-of-the-art.

Our review also found that few research papers provide a ready-to-use implementation of their work, making it easier for other researchers to benchmark their algorithms against. On the other hand, the ones that offer an implementation are commonly used by others as part of their evaluation. For instance, work by \citet{ten2017grasp} is used by many others as a benchmark, likely because it is one of the earliest works that provide an open-source implementation wrapped in a ROS package.

\textbf{Key Takeaways:}
\begin{itemize}
    \item 6-DoF grasping should be studied in more varied environments rather than just tabletop scenarios.
    \item Researchers should make their algorithms publicly available, ideally in a ready-to-use format (e.g. as a ROS package) to allow for informative benchmarking even on tabletop scenarios.
\end{itemize}

\subsection{Performance Metrics}

\citet{Bohg_2014} highlighted in 2014 that the grasping community has not yet embraced a consistent set of performance metrics. This observation is still valid today. This can be partly attributed to the large variety of objects, robots, end-effectors, and scenarios used in grasping research. Moreover, there is a divide between subsystem metrics and task-level metrics~\cite{Leitner2017TheAP}. 

There are, however, some common performance metrics that the community has been using. The most popular metric is the grasp success rate, even though the exact definition varies in the literature. Some deem a grasp successful if the object is still held by the gripper after the robot returns to a configuration that is a certain height above the table~\cite{wu2020grasp, varley2015generating}. Other works impose additional constraints to the success definition, such as if the object is held for a certain amount of time~\cite{choi2018soft} or checking if the object is still in the gripper after taking a sequence of actions intended to test the robustness of the grasp~\cite{lundell2020beyond}.






Although success rate is a useful metric, an underutilized type of performance metric is one that measures the time efficacy of a grasping approach. For instance, if the task is to remove objects from a container efficiently, is a slower robot with a 95\% success rate preferred to a faster robot with only 75\% success rate? The answer depends on the task, since faster grasping approach might be preferred for a task where dropping an object is not detrimental. One metric that could be used for this purpose is Mean Picks Per Hour (MPPH), which is defined as the average number of successful grasps completed in an hour. Researchers should be reminded, however, that a time-based metric such as MPPH does not only measure the computational efficiency of grasp synthesis but the system as a whole, and will be affected by the robot hardware, trajectory efficiency as well as the compute resources. 

The \emph{combination} of employing both a time-based metric and success rate could provide a more comprehensive evaluation of a pure grasping system. This is due to the likely inverse nature of the two metrics, where a system with a high grasp success rate may be very computationally expensive and slow-moving. However, a very fast system may be able to perform grasps and subsequent motions quickly, however likely decreasing the grasp success, which may be undesirable when grasping more fragile objects. As such, considering both of these metrics should provide a more detailed evaluation of the grasping system.






\textbf{Key Takeaways:}
\begin{itemize}
    \item In addition to reporting the grasp success rate, researchers should consider reporting a more strict definition of the grasp success that tests the robustness of the grasp, similar to \citet{lundell2020beyond}.
    \item We suggest wider adoption of a time-based performance metric, such as the Mean Picks Per Hour (MPPH).
\end{itemize}

\subsection{Object Sets and Grasping Datasets}

Many papers use custom object sets, consisting of daily objects that the authors could find around the lab, making it difficult to compare different grasping approaches. Standardized object sets are very useful in enabling head-to-head comparisons of grasp synthesis algorithms. The most useful object sets for grasping research are those which cover a variety of objects and are easily accessible by the research community. Furthermore, object sets should have accurate 3D models that enable simulation studies. The most commonly used object set currently by today's grasping community is the YCB object set~\cite{calli2015ycb}. However, since this may change in the future, researchers should keep an eye on which object sets the community is using.

Models for grasp synthesis are trained on datasets where each data point contains the sensor data, the grasp pose executed by the robot, along with a ground truth label depending whether the grasp was successful or not. The majority of the reviewed works were trained on simulated datasets which offer large-scale data collection. However, there is a lack of realism due to the physics not being modeled accurately~\citep{RUBERT2019103274}.There is a lack of real-world grasping datasets, which tend to  offer higher-quality data, are resource intensive to create compared to simulated datasets. While we believe that a comprehensive real-world dataset could be useful for the community analogous to the commonly used Cornell Grasping dataset~\cite{lenz2014deep} for 4-DoF top-down grasping, however, most research groups typically lack the time and resources needed to create large-scale real-world grasping datasets, with notable exceptions~\cite{levine2016learning}. A compromise between real and simulated datasets are the hybrid datasets, for instance, \citet{fang2020graspnet} whom provided real-world point cloud data, however the grasp hypotheses are evaluated analytically rather than with actual trial-and-error.

\textbf{Key Takeaways}: 
\begin{itemize}
\item Grasping researchers should adopt one of the widely used object sets. YCB object set~\cite{calli2015ycb} is the most commonly used one today, however, in the future, new object sets might find a wider adoption.
\item The grasping community would benefit from the introduction of new object datasets that offer variety or specialization in different aspects such as object geometry, grasping difficulty \cite{morrison2020egad} or deformation.
\item We propose authors release the code used to generate the dataset in addition to the dataset itself. This will allow other authors to make changes to the data generation procedure.
\end{itemize}

\subsection{Trajectory Planning}





Motion planning is widely utilized to find collision-free trajectories to execute a chosen grasp hypothesis. However, planning for a collision-free trajectory can be too conservative for densely cluttered scene, for example in cabinets and shelves where it may not possible to grasp the object of interest without nudging the neighboring objects. It has been shown that humans make deliberate contact with the environment during manipulation rather than carefully avoiding it\cite{deimel2016exploitation}. This strategy exploits how the environment provides physical constraints on how the object and hand move and therefore provides a funnel for uncertainty due to noise in perception and control. This principle has also been shown to work well for robots~\citep{deimel2016exploitation,kazemi2012robust,righetti2014autonomous,hudson2012end,toussaint2014dual,dafle2014extrinsic,chavan2015prehensile,shao2020learning}. For example, fixturing is a widely used practice in industry for various application such as machining, assembly and inspection. This suggests that the objective of having collision-free trajectories to acquire a grasp must be relaxed.

\textbf{Key Takeaway}: We encourage the study of motion planning algorithms for tight spaces where it is not possible to reach the targets without nudging other objects or where contact with the environment can be leveraged for more robust grasping.

\subsection{Sensor Modalities}

While most of the reviewed papers rely solely on vision for perceiving the world, other sensing modalities have been shown to be very helpful for both manipulation~\cite{doi:10.1177/17298806221095974} and grasping~\citep{9212350}. Future research should explore how these techniques can be applied to improve the success of 6-DoF grasping. For example, tactile sensors can be used to predict if the object will remain grasped before it is actually lifted~\cite{2011_TouchStability, 2018_MoreThanAFeeling, grasp_tactile}, detect object slip~\cite{SlipReview}, account for uncertainty in object pose~\cite{2019_ICRA_mbkrb} and reconstruction of object geometry~\citep{2021_ShapeStability_RAL, ottenhaus2019visuo}. Force/Torque sensing is another common sensor modality utilized in robotics there are also other less common modalities include robotic skins~\cite{hughes2018robotic} and sound~\cite{zoller2020active, gandhi2020swoosh}. 




\textbf{Key Takeaway}: Many of the reviewed works are vision-only sensing and considering only a single point-of-view. More research is needed to complement vision with other senses such as touch or hearing.


\subsection{Grippers, hands and beyond}
Most of the works in this survey focus on the use of simple two-fingered or three-fingered grippers, with few consider anthropomorphic hands. Simpler grippers greatly reduce the complexity of computing many-Dofs grasp hypotheses, and, most importantly, affordable commercial grippers allow researchers to setup systems for experimentally validating their results. However, this biases the research towards problems suitable for these simpler grippers and makes research on some other problems like grasping with anthropomorphic hand, dexterous, in-hand manipulation or soft-hand grasping fall behind. These directions may become important for enabling robots to go beyond pick and place tasks.

\textbf{Key Takeaway}: Robots have been employed and productized in manufacturing for decades~\citep{weldingRobot, EDWARDS198445, cho_warnecke_gweon_1987}, with further improvements in manipulation capabilities these robots will be able to perform more complex tasks in more unstructured environments, like the assembly of unseen parts or on-site construction. Building robust and dexterous hands will result in automating many jobs currently done by humans. The opportunities are immense: new ways of designing hands, new materials, new sensors and ways of actuation, easy exchangeable fingers, redundancy. But it is not only about the hands - making sure the arm can carry the hand, that the hand is a natural extension of the arm, hand and arms in interaction- going beyond one and compensating some degrees of freedom in the hand by skillful dual-arm interaction are just some of the avenues to follow.

\subsection{Grasping as part of a process}
All day long, our hands and fingers, touch, push, pull, and enclose objects. We do this with rigid, articulated or deformable objects. We do this in the air, while objects are standing on rigid or soft surfaces or while they are moving. We do this while they are in water or covered in oil, with or without seeing - like finding a key in a pocket. Despite many decades of research and development in the area, human grasping skills are still quite superior to any of the artificial systems so far demonstrated. 

In this discussion section, we have identified several areas where further contributions are needed. However, a broader view is necessary to address some if not all the manipulation scenarios mentioned above and, more importantly, consider grasping as part of an entire manipulation process. Specifically, identifying one specific 6DoF grasp pose for an object may not be sufficient to achieve success in these scenarios. One avenue may be to re-think grasp parameterizations to extend them to trajectories defined in joint and/or contact space. This could include exploiting environmental constraints~\citep{deimel2016exploitation} as well and adapting wrist trajectories based on the morphology of the hand itself~\citep{Gilday2021}. New ideas and techniques for online learning and recovering from failure are required. Only new ideas and techniques in these and more areas will push advances that will make robotics systems successfully perform tasks we have been promising for some time such as folding clothing, preparing food, dressing humans. Pushing for and studying grasping beyond pure grasping, is one of the most important challenge we have as a community. 

In summary, we have conducted a systematic review of state-of-the-art works which use deep learning-based towards achieving 6-DoF grasping. From this review, we synthesize 10 key takeaways which we believe could enable further progress in this rapidly progressing field.

\section{Acknowledgments}
We would like to thank Wesley Chan for the helpful discussion at the ideation stages of the work. We would like to thank Lily Tran for helping with table creating and editing.

\normalsize
\renewcommand*{\bibfont}{\footnotesize}
\bibliographystyle{myieee}
\bibliography{IEEEabrv,refs}

\section{Appendix}
\subsection{Appendix A: Methodology}
\footnotesize
\label{sec:method}

We conducted a Systematic Literature Review to assess the state-of-the-art of 6-DoF robotic grasping utilizing deep learning approaches. Our review, methodology adapted from~\cite{Shishegar2018}, searched through six different scholarly libraries -- IEEE, Springer, ScienceDirect, SpringerLink, arXiv and Taylor \& Francis -- with the following search terms and criteria.

\subsection{Search Terms}

The following search terms were used to search through the whole paper (where the database allowed this option). The `AND' function represents a way to enforce the search results to contain all of the search terms. When possible, the search terms were combined using an `OR` function,  otherwise, each term was searched separately. Furthermore, in combination with the search term `Grasping' was searched in the metadata of the paper, if metadata search was available.

\begin{itemize}
    \item 6-DoF Grasping
    \item Grasping AND Point Cloud AND Deep Learning
    \item Shape Completion AND 6-DoF Grasping AND Learning
    \item Affordances AND Grasping AND 6 DoF
\end{itemize}

\subsection{Inclusion/Exclusion Criteria}

From the papers found in the databases, only publications that met the following criteria are included in this review:
\begin{enumerate}[label=(\alph*)]
\item Paper considered grasping from a table-top scenario,
\item All 6-DoF were used for the grasp pose, 
\item Deep Learning methods were applied in some aspect of the work, 
\item Published after Jan 1, 2012, (the year Alexnet~\cite{krizhevsky2012imagenet} was published) and
\item Written in English.
\end{enumerate}


All criteria was searched manually by the authors. Our search returned a total of 85 papers that matched these criteria and were therefore included in this review. A table of all the reviewed papers and their deep-learning approaches are included in the online version of the survey.

\subsection{Data Extraction/Analysis}
Data was extracted from the included papers and verified by at least two of the authors.






\vspace{-25mm}
\begin{IEEEbiography}
    [{\includegraphics[width=1in,height=1.25in,clip,keepaspectratio]{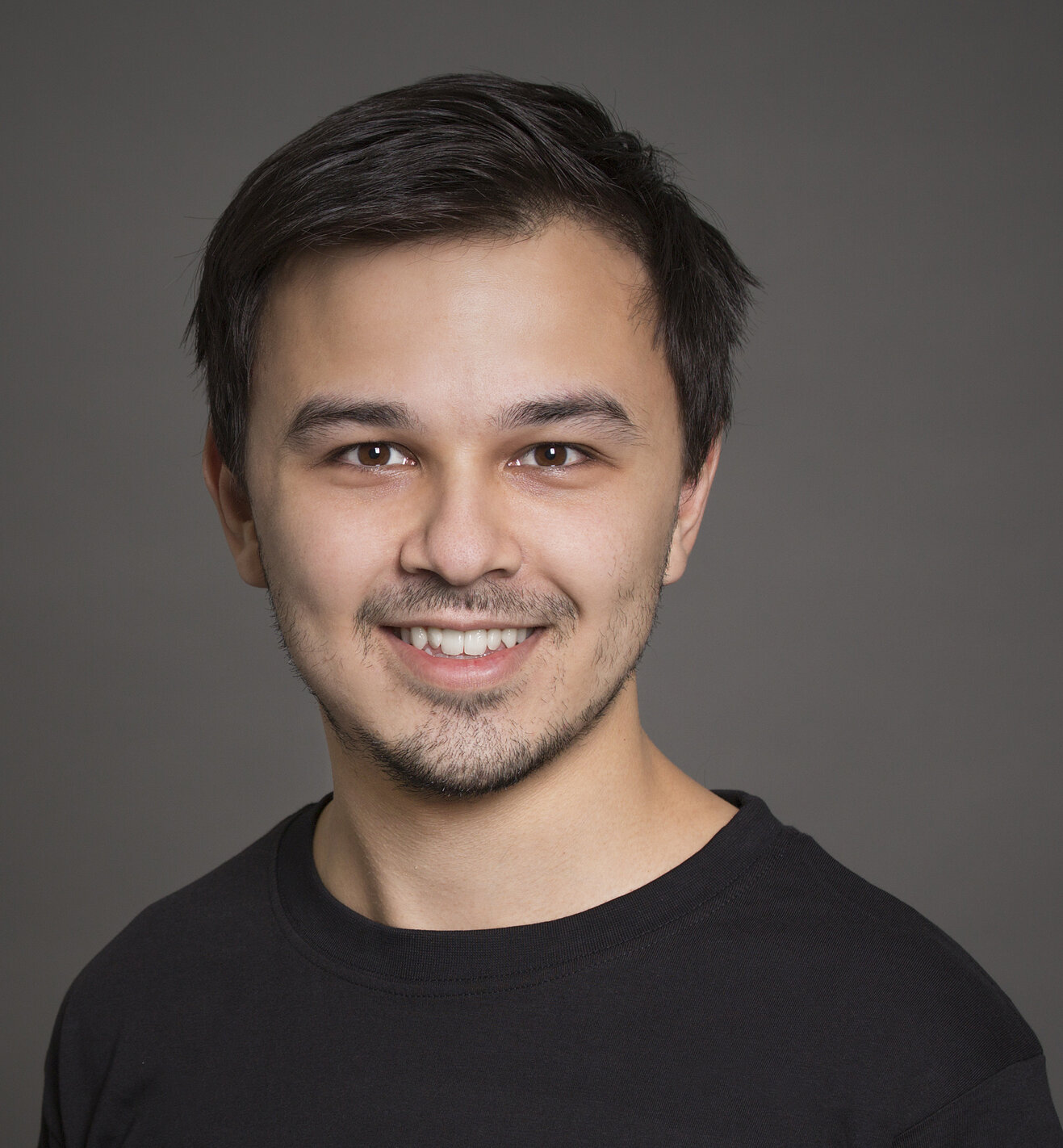}}]{Rhys Newbury}
is a PhD student at Monash University, Australia. He holds a B. Eng and B. Sc (Computer Science) from Monash University. His research interests focus around manipulation and human-robot interaction. He has a focus on systems view to problems and how to apply robots to real-world issues.
\end{IEEEbiography}
\vskip -2\baselineskip plus -1fil
\begin{IEEEbiography}
    [{\includegraphics[width=1in,height=1.25in,clip,keepaspectratio]{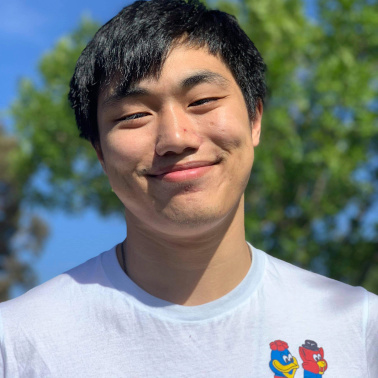}}]{Morris Gu}
is a PhD student at Monash University, Australia. He holds a B. Eng from Monash University. His research interests are in explainability and augmented reality within human-robot interaction. 
\end{IEEEbiography}
\vskip -2\baselineskip plus -1fil
\begin{IEEEbiography}
    [{\includegraphics[width=1in,height=1.25in,clip,keepaspectratio]{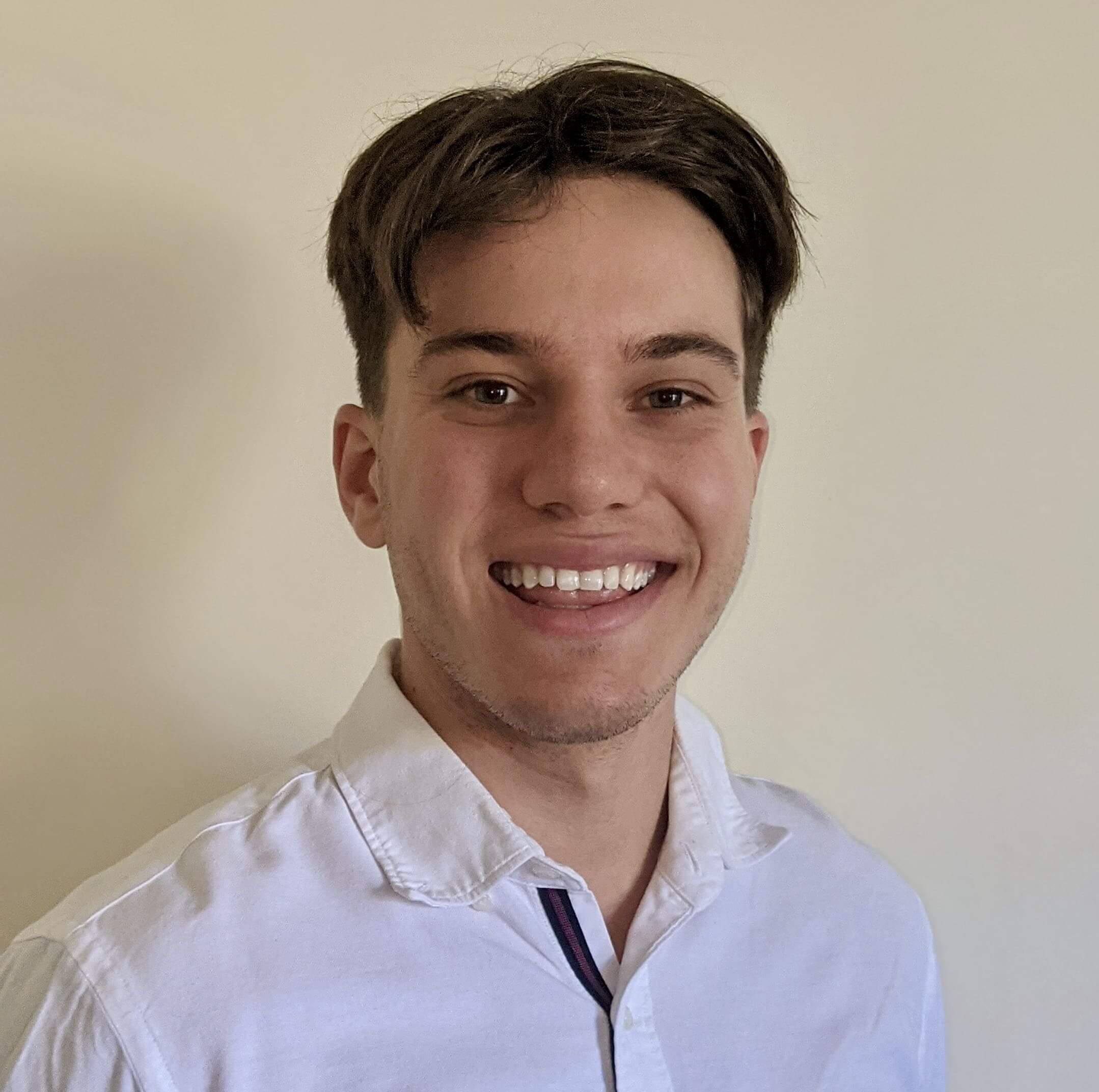}}]{Lachlan Chumbley}
is an undergraduate student at Monash University, Australia. His research interests include robotic grasping and manipulation, particularly though the use of tactile information.\end{IEEEbiography}
\vskip -2\baselineskip plus -1fil
\begin{IEEEbiography}
    [{\includegraphics[width=1in,height=1.25in,clip,keepaspectratio]{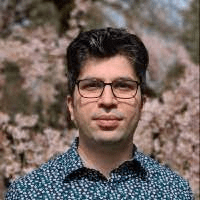}}]{Arsalan Mousavian}
received the B.Sc. degree from the Iran University of Science and Technology, Tehran, Iran, in 2010, the M.Sc. degree from the University of Tehran, Tehran, in 2013, both in AI and robotics, and the Ph.D. degree in computer science from George Mason University, Fairfax, VA, USA, in 2018. He is currently a Senior Research Scientist with NVIDIA, Seattle, WA, USA. His research interests include 3-D perception methods that help robots accomplish robot manipulation tasks in the real world.
\end{IEEEbiography}
\vskip -2\baselineskip plus -1fil
\begin{IEEEbiography}
    [{\includegraphics[width=1in,height=1.25in,clip,keepaspectratio]{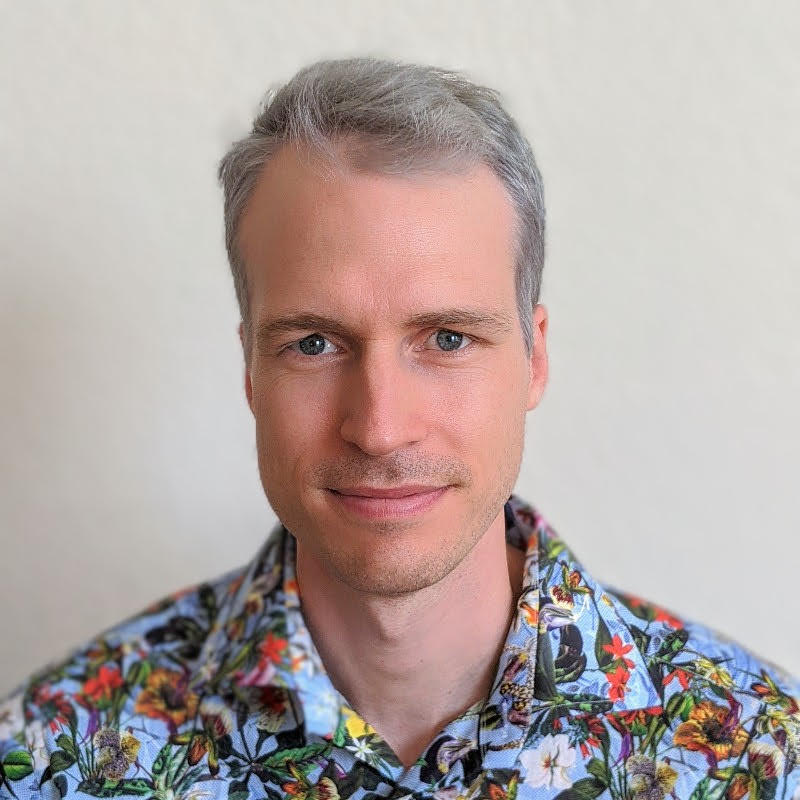}}]{Clemens Eppner} is a Research Scientist in the Seattle Robotics Lab at NVIDIA Research. Before joining NVIDIA, he received his Ph.D. at the Robotics and Biology Lab at Technische Universität Berlin, Germany, and M.Sc. from the University of Freiburg, Germany. He is interested in the problem space of grasping and manipulation, including aspects of planning, control, and perception.
\end{IEEEbiography}
\vskip -2\baselineskip plus -1fil
\begin{IEEEbiography}
    [{\includegraphics[width=1in,height=1.25in,clip,keepaspectratio]{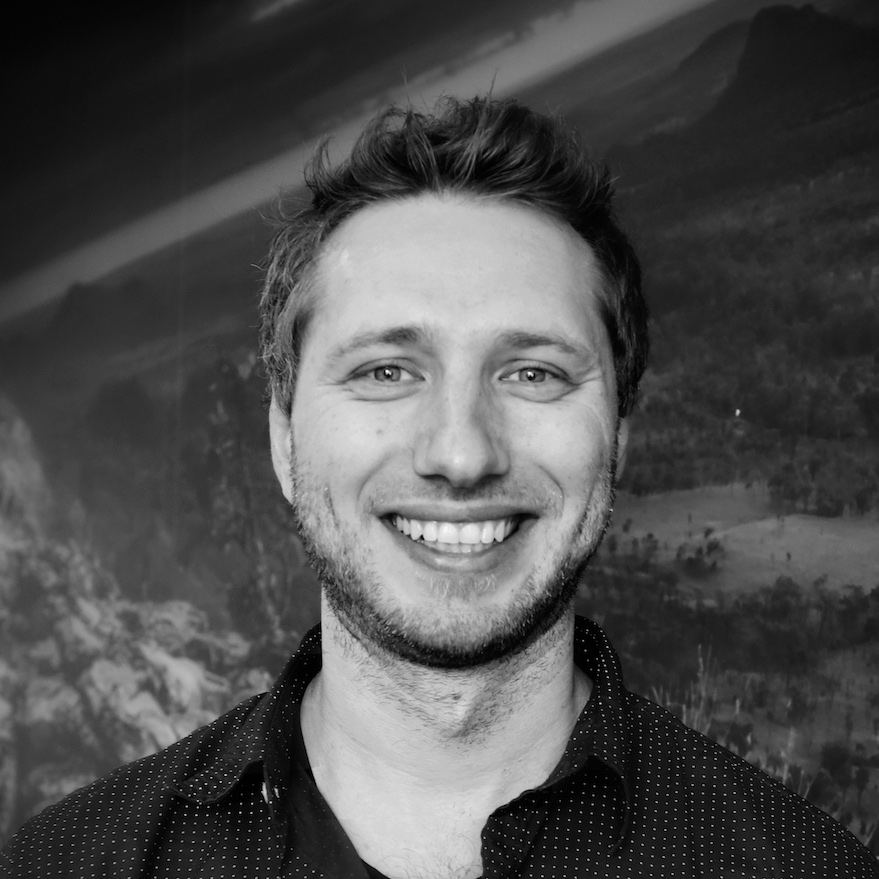}}]{J\"{u}rgen Leitner}
is co-founder at LYRO Robotics, Australia. He has designed, developed, and deployed robots fore more than 15 years. His interest lies in building intelligent robots that can safely and reliably interact with the physical world. He holds a PhD (2014), a MSc (Space Science, 2009), a MSc(Tech) (Space Robotics, 2009) and a BSc (Software Engineering, 2007).
\end{IEEEbiography}
\vskip -2\baselineskip plus -1fil
\begin{IEEEbiography}
    [{\includegraphics[width=1in,height=1.25in,clip,keepaspectratio]{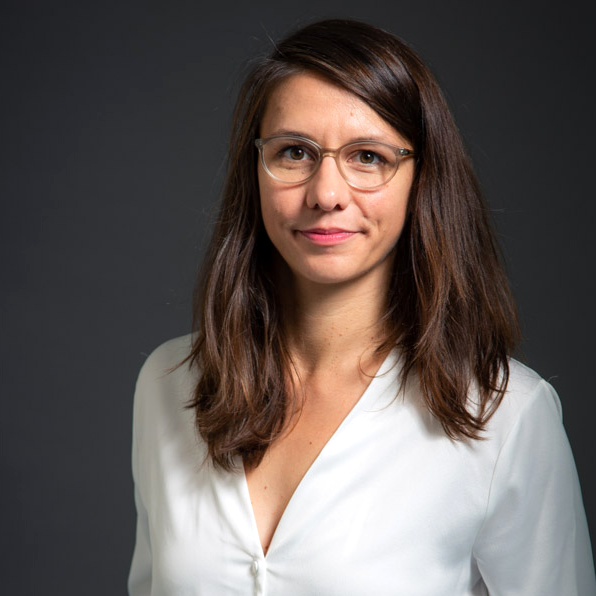}}]{Jeannette Bohg} is an Assistant Professor of Computer Science at Stanford University. She was a group leader at the Autonomous Motion Department (AMD) of the MPI for Intelligent Systems until September 2017. Before joining AMD in January 2012, Jeannette Bohg was a PhD student at the Division of Robotics, Perception and Learning (RPL) at KTH in Stockholm. Her research focuses on perception and learning for autonomous robotic manipulation and grasping.  
\end{IEEEbiography}
\vskip -2\baselineskip plus -1fil
\begin{IEEEbiography}
    [{\includegraphics[width=1in,height=1.25in,clip,keepaspectratio]{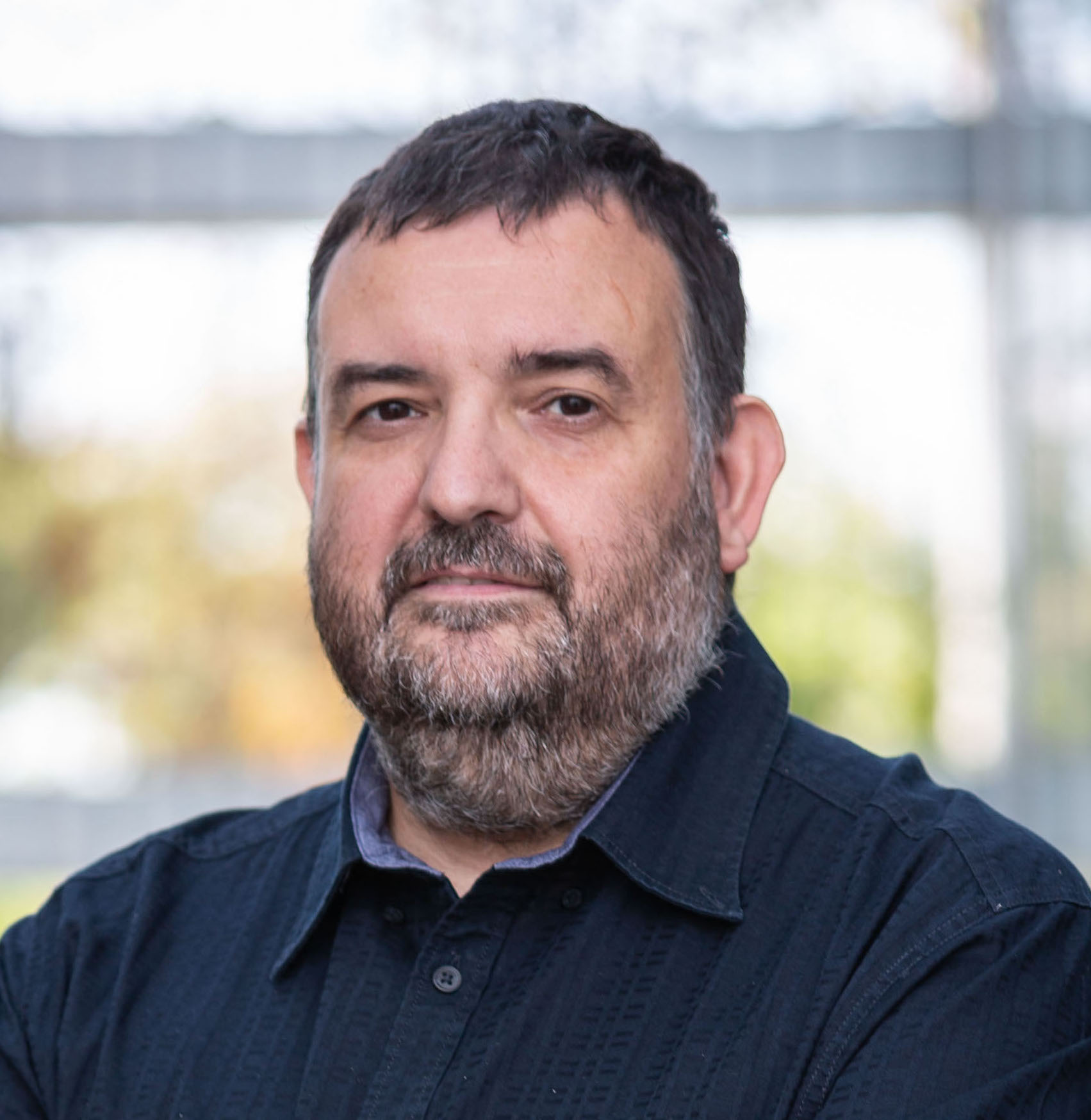}}]{Antonio Morales}
is Associate Professor at the Department of Computer Engineering and Science in the Universitat Jaume I of Castelló, Spain. He received his PhD in Computer Science Engineering from Universitat Jaume I in January 2004. Currently, he is director of the Degree Program on Robotics Intelligence and vice-dean of the School on Technology and Experimental Sciences of the Universitat Jaume I. He is a leading researcher at the Robotic Intelligence Laboratory at Universitat Jaume I and his research interests are focused on reactive robot grasping and manipulation. He has been a Principal Investigator on several European and national research projects.
\end{IEEEbiography}
\vskip -2\baselineskip plus -1fil
\begin{IEEEbiography}
    [{\includegraphics[width=1in,height=1.25in,clip,keepaspectratio]{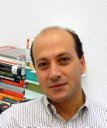}}]{Tamim Asfour}
is Professor at the Institute for Anthropomatics and Robotics at KIT, where he holds the chair of Humanoid Robotics Systems and is head of the High Performance Humanoid Technologies Lab (H$^2$T). His research interest is 24/7 humanoid robotics. Specifically, he studies the mechano-informatics of humanoids as the synergetic integration of informatics, artificial intelligence, and mechatronics into complete humanoid robot systems, which are able to perform versatile tasks real world. He is developer and the leader of the development team of the ARMAR humanoid robot family.
\end{IEEEbiography}
\vskip -2\baselineskip plus -1fil
\begin{IEEEbiography}
    [{\includegraphics[width=1in,height=1.25in,clip,keepaspectratio]{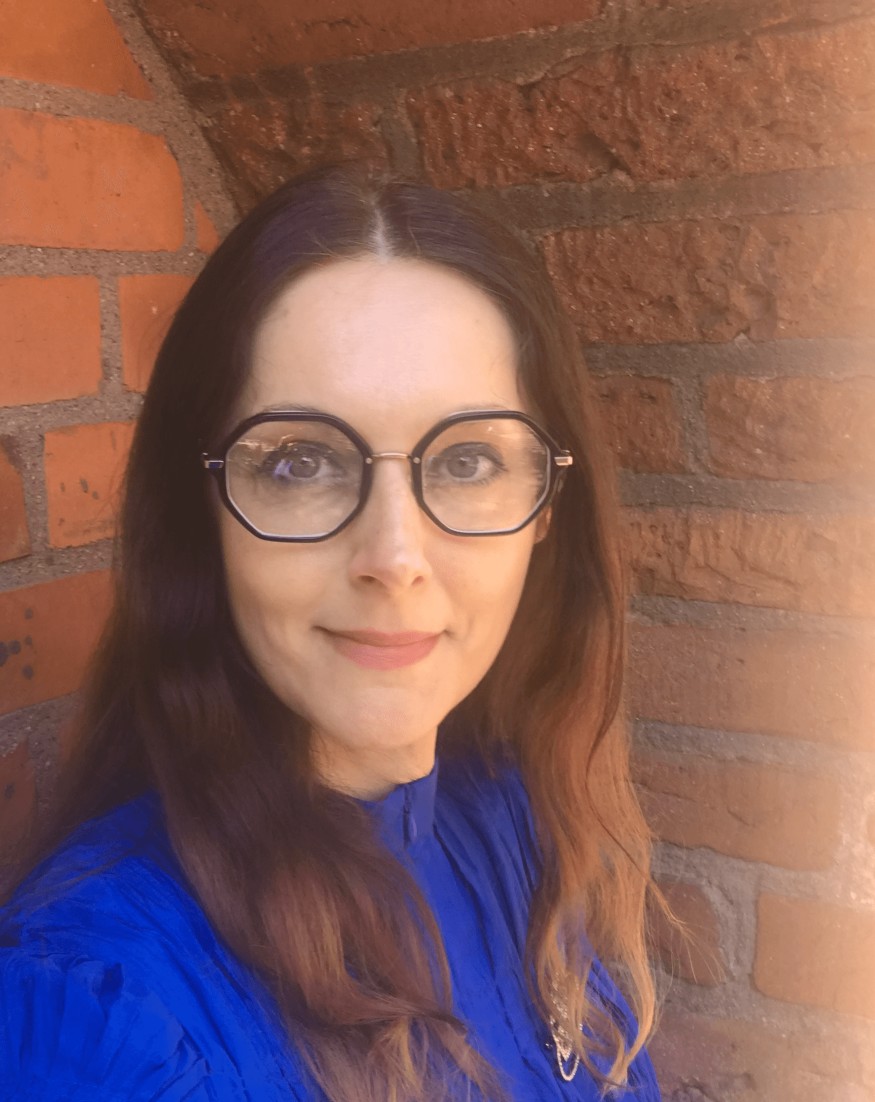}}]{Danica Kragic} 
is a Professor at the School of Computer Science and Communication at KTH in Stockholm. She received MSc in Mechanical Engineering from the Technical University of Rijeka, Croatia in 1995 and PhD in Computer Science from KTH in 2001. Danica received the 2007 IEEE Robotics and Automation Society Early Academic Career Award. She is a member of the Swedish Royal Academy of Sciences and Swedish Young Academy. She has chaired the IEEE RAS Technical Committee on Computer and Robot Vision and from 2009 serves as an IEEE RAS AdCom member.
\end{IEEEbiography}
\vskip -2\baselineskip plus -1fil
\begin{IEEEbiography}
    [{\includegraphics[width=1in,height=1.25in,clip,keepaspectratio]{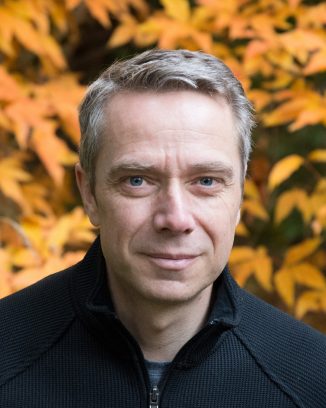}}]{Dieter Fox}
(Fellow, IEEE) received the Ph.D. degree in computer science from the University of Bonn, Bonn, Germany, in 1998. He is currently a Professor with the Allen School of Computer Science and Engineering, University of Washington, Seattle, WA, USA, where he heads the UW Robotics and State Estimation Lab. He is also Senior Director of Robotics Research with NVIDIA.He has authored or coauthored more than 200 technical papers. He is a co-author of the textbook entitled Probabilistic Robotics (Cambridge, MA, USA: MIT Press, 2005). His research interests include robotics and artificial intelligence, with a focus on state estimation and perception applied to problems such as mapping, object detection and tracking, manipulation, and activity recognition.
\end{IEEEbiography}
\vskip -2\baselineskip plus -1fil
\begin{IEEEbiography}
    [{\includegraphics[width=1in,height=1.25in,clip,keepaspectratio]{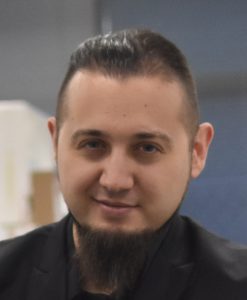}}]{Akansel Cosgun} 
is currently a Senior Lecturer at Deakin University, Australia. He received his Ph.D. degree in robotics from the Georgia Institute of Technology, Atlanta, GA, USA, in 2016. From 2018 to 2022, he was a Research Fellow with Monash University, Australia. He conducts research in robotics, human–robot interaction, and robot learning. From 2019 to 2020, he was the Team Lead for Vision-Based Manipulation with the Australian Centre for Robotic Vision. He has previously worked with Honda Research, Toyota Infotechnology Center, Microsoft Research, and Savioke, a robotics start-up. His research interests include mobile robots, robotic arms, and self-driving cars with an emphasis on a systems view to problems.
\end{IEEEbiography}

\end{document}